\documentclass[10pt,journal,compsoc]{IEEEtran}

\ifCLASSOPTIONcompsoc
  \usepackage[nocompress]{cite}
\else
  \usepackage{cite}
\fi

\ifCLASSINFOpdf
\else
\fi

\usepackage{amsmath,amssymb} 
\usepackage{color}
 \newcommand\figcaption{\def\@captype{figure}\caption}
 \newcommand\tabcaption{\def\@captype{table}\caption}
 \usepackage{subcaption}
\usepackage{graphicx}
\usepackage{amsmath}
\usepackage{amssymb}
\usepackage{booktabs}
\usepackage{textcomp}  
\usepackage{amsfonts}
\usepackage{multirow}
\usepackage{algorithm}
\usepackage{algorithmic}
\usepackage{float}
\usepackage{tabularx}
\usepackage{array}
\usepackage[misc]{ifsym}
\usepackage[pagebackref=true,breaklinks=true,colorlinks,bookmarks=false]{hyperref}

\usepackage{xspace}

\usepackage[pagebackref=true,breaklinks=true,colorlinks,bookmarks=false]{hyperref}

\def \etal{\emph{et al.}}
\newcommand*{\eg}{\emph{e.g.}}
\newcommand*{\ie}{\emph{i.e.}}

\usepackage{ragged2e}

\usepackage{overpic}
\usepackage{enumitem} 
\usepackage{overpic} 
\usepackage{color}

\definecolor{turquoise}{cmyk}{0.65,0,0.1,0.3}
\definecolor{purple}{rgb}{0.65,0,0.65}
\definecolor{dark_green}{rgb}{0, 0.5, 0}
\definecolor{orange}{rgb}{0.8, 0.6, 0.2}
\definecolor{red}{rgb}{0.8, 0.2, 0.2}
\definecolor{darkred}{rgb}{0.6, 0.1, 0.05}
\definecolor{blueish}{rgb}{0.0, 0.3, .6}
\definecolor{light_gray}{rgb}{0.7, 0.7, .7}
\definecolor{pink}{rgb}{1, 0, 1}
\definecolor{greyblue}{rgb}{0.25, 0.25, 1}

\newcommand{\Fig}[1]{Fig.~\ref{fig:#1}}

\newcommand{\Table}[1]{Table~\ref{tab:#1}}
\newcommand{\eq}[1]{(\ref{eq:#1})}

\newcommand{\Sec}[1]{Sec.~\ref{sec:#1}}

\usepackage{blindtext}

\renewcommand{\paragraph}[1]{\vspace{1em}\noindent\textbf{#1}.}

\usepackage{multirow}
\usepackage{amssymb}
\usepackage{subcaption}
\newcommand{\nn}{MCTformer}

\def \etal{{et al.}}

\begin{document}
%
\title{MCTformer+: Multi-Class Token Transformer for Weakly Supervised Semantic Segmentation}

%
%
%
%
\author{
Lian Xu, Mohammed Bennamoun, Farid Boussaid, Hamid Laga, Wanli Ouyang, Dan Xu,~\IEEEmembership{Member,~IEEE}

\IEEEcompsocitemizethanks{
    \IEEEcompsocthanksitem Lian Xu, Mohammed Bennamoun are with the Department of Computer Science and Software Engineering, The University of Western Australia (UWA), Perth. \protect~E-mail: \{lian.xu, mohammed.bennamoun\}@uwa.edu.au
    \IEEEcompsocthanksitem Farid Boussaid is with the Department of Electrical, Electronic and Computer Engineering, The University of Western Australia (UWA), Perth. \protect~E-mail: farid.boussaid@uwa.edu.au
    \IEEEcompsocthanksitem Hamid Laga is with the School of Information Technology, Murdoch University, Perth. \protect~E-email: H.Laga@murdoch.edu.au
    \IEEEcompsocthanksitem Wanli Ouyang is with Shanghai AI Laboratory, Shanghai. \protect~E-email: wanli.ouyang@sydney.edu.au
    \IEEEcompsocthanksitem Dan Xu is with the Department of Computer Science and Engineering, The Hong Kong University of Science and Technology (HKUST), Hong Kong.\protect~E-mail: danxu@cse.ust.hk}
}

%
%

\markboth{Journal of \LaTeX\ Class Files,~Vol.~14, No.~8, August~2015}
{Shell \MakeLowercase{\etal}: Bare Demo of IEEEtran.cls for Computer Society Journals}

\IEEEtitleabstractindextext{%
\justifying
\begin{abstract}
This paper proposes a novel transformer-based framework that aims to enhance weakly supervised semantic segmentation (WSSS) by generating accurate class-specific object localization maps as pseudo labels. 
Building upon the observation that the attended regions of the one-class token in the standard vision transformer can contribute to a class-agnostic localization map, we explore the potential of the transformer model to capture class-specific attention for class-discriminative object localization by learning multiple class tokens. 
We introduce a Multi-Class Token transformer, 
which incorporates multiple class tokens to enable class-aware interactions with the patch tokens. 
To achieve this, we devise a class-aware training strategy that establishes a one-to-one correspondence between the output class tokens and the ground-truth class labels. Moreover, a Contrastive-Class-Token (CCT) module is proposed to enhance the learning of discriminative class tokens, enabling the model to better capture the unique characteristics and properties of each class. 
As a result, class-discriminative object localization maps can be effectively generated by leveraging the class-to-patch attentions associated with different class tokens.
To further refine these localization maps, we propose the utilization of patch-level pairwise affinity derived from the patch-to-patch transformer attention. 
Furthermore, the proposed framework seamlessly complements the Class Activation Mapping (CAM) method, resulting in significantly improved WSSS performance on the PASCAL VOC 2012 and MS COCO 2014 datasets.
These results underline the importance of the class token for WSSS. The codes and models are publicly available \href{https://github.com/xulianuwa/MCTformer}{here}.
\end{abstract}

\begin{IEEEkeywords}
Weakly supervised learning, Transformer, Class token, Semantic Segmentation, Object localization.
\end{IEEEkeywords}}

\maketitle

\IEEEdisplaynontitleabstractindextext

%
\IEEEpeerreviewmaketitle

\section{Introduction}
\IEEEPARstart{T}{raditional} semantic segmentation approaches typically rely on precisely annotated pixel-level labels, which are costly and label-intensive to acquire. Weakly Supervised Semantic Segmentation (WSSS) methods, on the other hand, operate under the premise of weak supervision, which involves utilizing more readily available and less precise forms of annotations. The primary goal of WSSS is to achieve accurate segmentation results while minimizing the need for pixel-level annotations. Instead, weakly supervised methods make use of weaker forms of supervision, including image tags~\cite{li2019guided, jiang2021online, wang2022looking, lee2023saliency}, scribbles~\cite{zhang2021affinity}, and bounding boxes~\cite{zhang2021affinity}. These weak annotations provide only limited information about the spatial extent of objects or classes within an image, without specifying the exact boundaries of each region.

An essential aspect of the WSSS task involves generating high-quality pseudo-semantic masks utilizing weak labels. This is generally achieved by relying on Class Activation Mapping (CAM)~\cite{zhou2016learning} with Convolutional Neural Networks (CNNs). The CAM technique, however, can only provide coarse and imprecise class-specific dense localization maps. To address this problem, WSSS techniques employ various algorithms and strategies, including leveraging additional cues and priors such as contextual information, co-occurrence statistics, or relationships between classes, to infer pixel-level segmentation masks from the weak annotations. While WSSS methods have made significant progress, they still face various challenges. The ambiguity and noise in weak labels, as well as the inherent problems of the CNN architecture (\textit{e.g.,} the limited receptive field), can lead to imperfect pseudo ground-truth semantic masks, requiring complex algorithms to handle uncertainty and improve accuracy.

\begin{figure}[t]
\begin{center}
\includegraphics[width=0.98\linewidth]{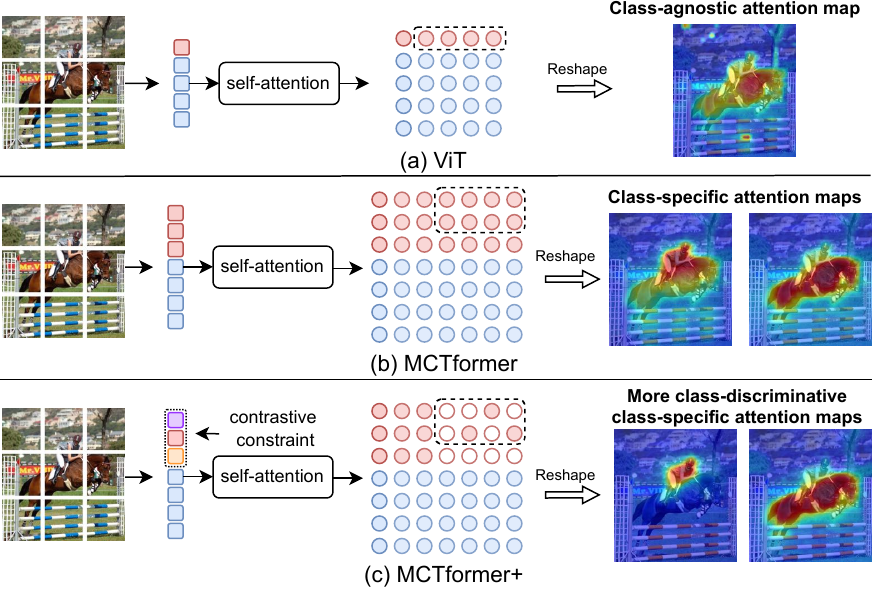}
\end{center}
\vspace{-8pt}
\caption{
(a) In the conventional ViTs~\cite{dosovitskiy2020image}, a single class token (red square) was utilized to consolidate patch token information (blue squares). Consequently, the resulting class-to-patch attention forms a single localization map independent of classes. (b) In the proposed \nn~\cite{xu2022multi}, multiple class tokens were learned to attend different patches. This resulted in diverse class-to-patch attentions, which can be utilized as localization maps of different object classes. (c) In the proposed \nn+, an additional class-contrastive constraint is imposed on the class tokens, leading to more class-discriminative localization.}
\vspace{-12pt}
\label{fig:teaser}
\end{figure}

The Vision Transformer (ViT), the pioneering transformer model tailored for computer vision, has exhibited remarkable performances in various visual tasks. Notably, ViT has achieved remarkable success in large-scale image recognition by harnessing its exceptional ability to capture extensive contextual information. ViT processes image data by partitioning an images into patches and
transforms them into a sequential representation of vectors. 
By leveraging the token-based design and incorporating the self-attention-based data processing mechanism, ViT facilitates the establishment of meaningful connections between patches, capturing dependencies and relationships within the image.
Moreover, a special feature of ViT lies in its utilization of \emph{one} additional class token, which consolidates information from the entire sequence of patch tokens. While a few transformer methods~\cite{chu2021twins,chu2021conditional, pan2021scalable} have omitted the class token, this work emphasizes its critical significance, particularly for WSSS. 
Caron~\etal~\cite{caron2021emerging} observed that the image features from a self-supervised ViT could explicitly present the semantic segmentation information. In particular, it was noticed that the class token attention enables the discovery of a semantic scene layout, yielding encouraging outcomes in the unsupervised segmentation task.
While Caron~\etal~\cite{caron2021emerging} has shown that various heads in the multi-head attention layer of ViT can focus on semantically distinct image regions, the method to accurately link a head to a semantic class remains uncertain. In other words, these attention maps are still independent of specific classes (see \Fig{teaser} (a)).

\par Exploiting class-specific attention from transformers poses a significance challenge. We argue that using a \emph{single} class token hinders transformers' ability in localizing various objects within a single image.
This arises from two primary reasons. 
(\textbf{i}) the presence of a single class token inherently enables the learning of a diverse mix of image information, encompassing various classes and background contexts.
Consequently, both class-specific and generic object features are captured in this single class token. This inevitably leads to a noisy and less class-discriminative localization.
(\textbf{ii}) one class token lacks the capacity to effectively model the complex relationships with patch tokens for multiple diverse categories within a dataset, resulting in imprecise localization of different objects.

To address these limitations, a simple solution involves utilizing multiple class tokens, each targeted for learning representations of a specific class.
To achieve this, we proposed a Multi-Class Token transformer (\nn)~\cite{xu2022multi}, 
which employs \emph{multiple} class tokens to capture unique relationships with different objects through transformer attention. However, a mere increase in the number of class tokens in ViT does not inherently assign them specific meanings. 
In order to ensure the discriminative ability of individual class tokens for their respective object categories, a class-aware training strategy was proposed. 
More specifically, we process the produced multiple class tokens by the final layer of the transformer encoder by performing class-wise average pooling. This generates class scores that receive direct supervision from the ground-truth image-level labels, forming a one-to-one mapping from each class token to its respective class label.
The inherent benefit of this design lies in the direct use of the learned attention between each class token and patch tokens as class-specific localization maps for different object categories.

It is important to highlight that the transformer attention learned between patch tokens inherently yields patch-level pairwise affinity without any additional computation during training. This valuable information can be leveraged to enhance the class-specific transformer attention maps, resulting in smoother and more coherent boundaries as well as improved continuity, 
significantly facilitating localization performance.
Furthermore, this work also demonstrates that the proposed~\nn~synergizes with CAM, resulting in a powerful combination when implemented on patch tokens (which we refer to PatchCAM). Through the joint learning of multiple class tokens and patch tokens towards a same classification objective, a strong alignment between them is achieved, greatly strengthening the class-discriminative capability of the resulting localization maps.

In this paper, we propose \nn+, which improves \nn~\cite{xu2022multi} primarily from two aspects:
(\textbf{i}) To enhance the class-discriminative ability of the class-to-patch transformer attention maps, we introduce additional regularization losses on the output class tokens. These regularization losses ensure that the class tokens are distinct from one another, compelling them to attend to different patch tokens. This encourages the model to learn more diverse and specific attention patterns for each class, leading to improved class discrimination and localization capabilities (see \Fig{teaser} (b) and (c)). (\textbf{ii}) We introduce the use of the global weighted ranking pooling to replace the global average pooling, for the aggregation of the output patch tokens to predict class scores. Different weights are assigned to the patch tokens based on their relevance to the target object class. This weighted pooling strategy ensures that more informative and discriminative patch tokens contribute more significantly to the final class score prediction. As a result, the PatchCAM maps exhibit a notable improvement, with a significantly reduced inclusion of unrelated background regions. This effectively enhances the precision and accuracy of object localization. 
These two methods collectively contribute to improved class-specific dense localization performance, thereby improving the overall WSSS performance.

The main contribution of this paper is three-fold:
\begin{itemize}
    \item We propose a novel multi-class token transformer (\nn+) for weakly supervised semantic segmentation. To the best of our knowledge, it is the first work that exploits transformer attention for the generation of class-specific localization maps.
    \item To enable the class-specific token learning, we propose a class-aware training strategy and a contrastive-class-token module, jointly contributing to the class-discriminative multi-class transformer attention maps.
    \item The proposed method fully exploits the transformer attention for WSSS. We extract the class-specific localization maps from the transformer attention between each class token and patch tokens. We also propose to use the transformer attention between patches as a patch-level pairwise affinity, greatly enhancing the generated localization maps.
\end{itemize}

The effectiveness of each component of the proposed \nn+ has been extensively validated. The proposed \nn+ compared against the latest advanced methods on the PASCAL VOC 2012 and MS COCO 2014 benchmarks for WSSS and on the OpenImages dataset for weakly supervised object localization, demonstrating the superiority of the proposed approach.

\section{Related Work}
\label{sec:related}

\subsection{Weakly supervised semantic segmentation} 
Class Activation Mapping (CAM)~\cite{zhou2016learning}, as a classic weakly supervised object localization method, has been widely adopted in the existing WSSS works. However, CAM is unable to generate complete object regions and accurate object boundaries. Extensive effort has been carried out to address this limitation through various means in prior research.
For example, specific segmentation loss functions~\cite{kolesnikov2016seed,tang2018regularized, zhang2019reliability,ke2021universal} have been proposed to handle the issue of deficient segmentation supervision. Many other works have focused on enhancing the CAM maps to provide high-quality supervision for semantic segmentation. We categorize these methods broadly into the following groups:

\par\noindent\textbf{High-quality CAM generation.} To enable the network to capture more object parts, a common solution involves introducing greater challenges to achieve the classification objective. This has been implemented through various modifications to either the input data~\cite{singh2017hide,wei2017object,lee2021anti,zhang2021complementary, li2018tell} or feature maps~\cite{lee2019ficklenet, hou2018self, choe2019attention}, including techniques like dropping out certain image information or introducing perturbations. Some works~\cite{chang2020weakly} have advanced the task to a more challenging classification objective by introducing more fine-grained categories. Additional information across multiple images has been exploited to enhance the CAM maps~\cite{fan2020cian, sun2020mining, li2021group}.
To address the limitations that standard image classification objective loss functions do not ensure the discovery of complete object regions, several studies have proposed regularization losses~\cite{wang2020self, zhang2020splitting} to encourage the network to attend to additional object parts.
Several other works~\cite{wei2018revisiting} have identified that the local receptive fields of conventional image-classification CNNs hinder the propagation of discriminative information and proposed to incorporate multi-scale dilated convolutions to obtain more complete localization maps.  
Due to the prevalence of pre-trained large-scale vision-language models \eg, CLIP in various visual tasks, several works~\cite{xie2022clims, lin2023clip, xu2023learning} have exploited the rich image-language context for generating more accurate CAM maps.

\par\noindent\textbf{Learning class representations.} The CAM mechanism can be interpreted as utilizing the pixel-wise association between the classifier's class-related weights and image features to produce localization maps specific to each class. These class-related weights can be seen as class representations or class centers. Prior works have identified that the deficiency of the CAM maps is also attributed to the inability of the class representations. Therefore, several works have focused on learning more robust and adaptive class representations. Image-Specific CAM (IS-CAM)~\etal~\cite{chen2022self} was proposed to learn image-specific prototypes to capture these pixel-wise semantic correlations. IS-CAM aggregates the structure-aware seed regions, where the seed regions are determined by the consistency between the CAM maps and the pair-wise pixel feature similarities. Du~\etal~\cite{du2022weakly} proposed to construct class prototypes by aggregating the pixel features whose corresponding CAM scores fall within the top K confidences. These prototypes help enhance the learning of discriminative dense visual representations by attracting pixels towards the prototypes for their respective classes while driving them further from the prototypes for other classes. The proposed multi-class token learning is closely related to class representation learning. Unlike those feature aggregation-based methods, the proposed method explicitly learns class representations by designing multiple class tokens. The transformer attention between each class token and patch tokens from different layers represent different semantic-level class-to-pixel correlations, jointly contributing to precise class-specific localization maps. 

\par\noindent\textbf{Affinity learning for CAM refinement.} Several studies have developed methods of learning pairwise semantic affinities for CAM map refinement. For instance, Ahn~\etal~\cite{ahn2018learning} introduced AffinityNet, a method that learns pixel-wise affinities between neighboring pixels using the CAM-derived affinity pseudo labels. The learned affinities are utilized to perform random walk, enabling CAM propagation. 
Likewise, Wang~\etal~\cite{wang2020weakly} developed an affinity-learning network utilizing affinity pseudo labels generated from the confident segmentation predictions. Some other works~\cite{wang2020self,zhang2021complementary} have exploited feature affinities from image classification networks. Multi-task feature affinities have also been investigated in a weakly supervised framework for saliency detection and semantic segmentation~\cite{xu2021leveraging}.
An Affinity from Attention (AFA)~\cite{ru2022learning} module was proposed to predict affinities based on the transformer attention between patches using the affinity pseudo labels derived from the semantic segmentation pseudo labels.
Lin~\etal~\cite{lin2023clip} proposed class-aware attention-based affinity, which applies class-specific masks indicating the reliable CAM regions on the transformer pairwise patch attention maps.

A notable aspect is that the proposed \nn~was the first work that exploited class-specific transformer attention for discriminative localization. The proposed \nn+ further improved the class-specific localization performance, achieving superior results compared to existing WSSS methods.

\begin{figure*}
\begin{center}
\includegraphics[width=.95\textwidth]{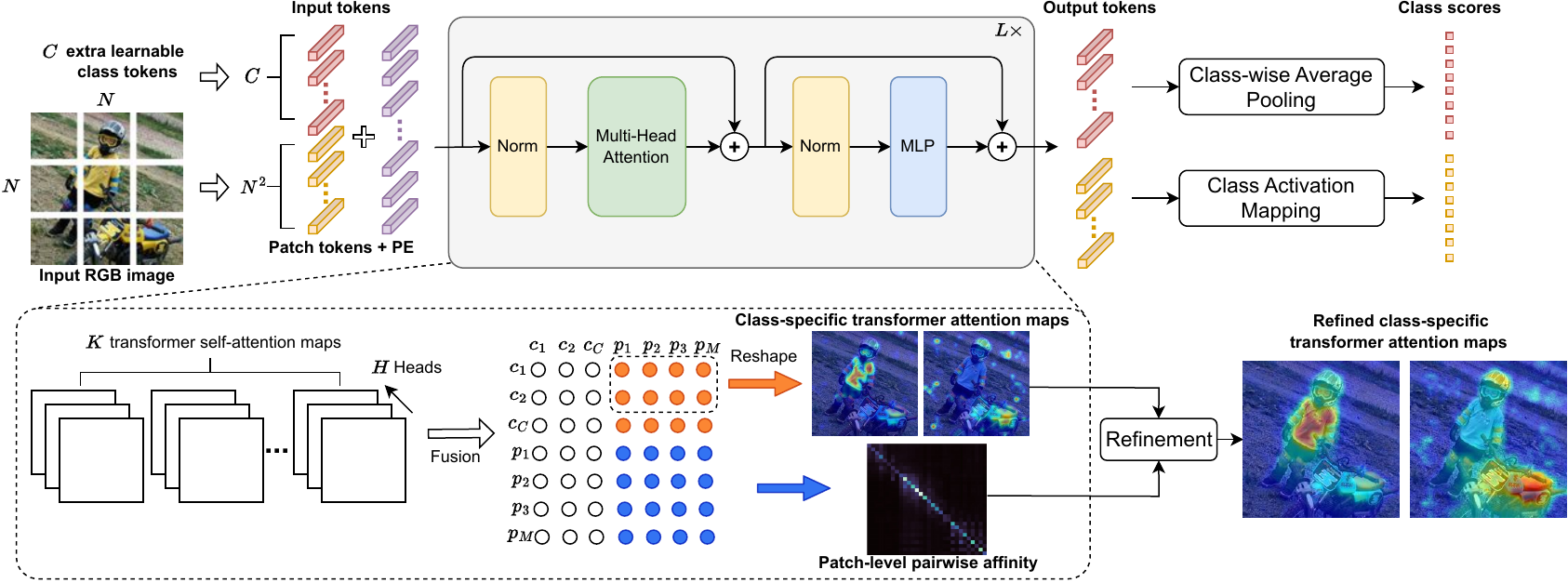}
\vspace{-8pt}
\end{center}
\caption{Illustration of the proposed~\nn. As input, an image is partitioned into patches, which are projected into patch tokens. We introduce $C$ additional class tokens, where $C$ denotes the number of classes. These class tokens are combined with patch tokens and augmented with positional embeddings (\textbf{PE}). The resulting sequence passes through $L$ consecutive transformer encoding layers. The output $C$ class tokens are utilized for generating class scores through class-wise average pooling, and the output patch tokens pass through a Class Activation Mapping (CAM) module (\Fig{cam}), producing class scores.
At inference time, the attention maps from multiple ($L$) transformer encoding layers are combined and averaged across multiple heads (H). From this fusion map, two types of attention maps can be extracted: (\textbf{i}) class-to-patch attention (yellow dots), which can be transformed into class-specific localization maps; (\textbf{ii}) patch-to-patch attention (blue dots), which acts as a pairwise affinity capable of refining the localization maps.
}
\label{fig:overview}
\vspace{-8pt}
\end{figure*} 

\vspace{-4pt}
\subsection{Visual applications of transformers}
Transformers~\cite{vaswani2017attention} were initially developed for processing sequential data in natural language processing tasks, they have more recently been applied on visual data for a diverse array of computer vision tasks~\cite{khan2022transformers, dosovitskiy2020image,liu2021visual, ranftl2021vision}, yielding promising performances. 
The pioneering vision model based on Transformers based is ViT~\cite{dosovitskiy2020image}, which operates on image patches. The self-attention module, the core component of ViT, allows each patch to interact with all other patches in the image, effectively providing a global receptive field for each patch.

Caron~\etal~\cite{caron2021emerging} has trained a self-supervised ViT, leading to an observation that the transformer attention between the class token and patch tokens effectively captures the layout information of scenes.
However, their discoveries were constrained to the unsupervised setting, where class labels were not assigned to the attention.
TS-CAM~\cite{gao2021ts} has integrated a CAM module to a ViT, enabling class-discriminative localization within ViT. This is however owing to CAM as ViT only provides the attention maps that are irrespective of classes in TS-CAM.
In contrast, the proposed \nn~exploits class-specific transformer attention for discriminative localization. This is shown to be a more effective complement to the CAM mechanism, resulting in superior class-specific localization compared to TS-CAM (see \Table{abla_mct_pgt}).

\par\noindent\textbf{ViTs for WSSS.} MCTformer~\cite{xu2022multi} and AFA~\cite{ru2022learning} are the first two concurrent works which used ViTs for WSSS. AFA~\cite{ru2022learning} focuses on leveraging the self-attention between patch tokens to predict semantic affinity, which can be used to refine the CAM maps. MCTformer~\cite{xu2022multi} introduced multiple class token learning to exploit class-specific localization information from the transformer attention between each class token and patch tokens. This led to a significant improvement in the refined localization maps when using the transformer attention between patches as affinities, requiring no extra supervision. Ru~\etal~\cite{ru2023token} proposed ToCo, which addressed the over-smoothing issue of the CAM maps derived from the top layer by enforcing the consistency of the CAM maps from different layers in terms of the pairwise patch token similarity. ToCo also introduced a contrastive loss between the class tokens of the global image and its local crops, enhancing the class token to attend the entire object regions.

\section{Multi-class Token Transformer}
\label{sec:methodology}
\subsection{Overview}
\begin{figure*}
\begin{center}
\includegraphics[width=.98\textwidth]{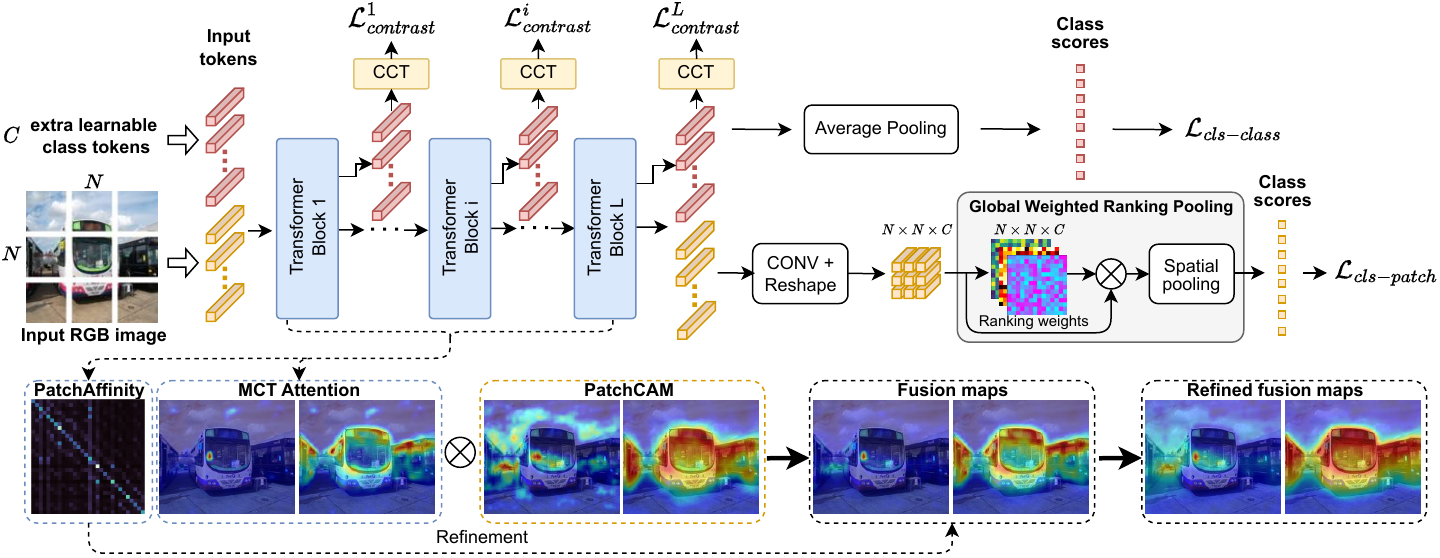}
\vspace{-8pt}
\end{center}
\caption{Illustration of the proposed \nn+. \nn+ proposes the CCT module (illustrated in~\Fig{cct}), which is applied on the output class tokens of each layer, to enhance the learning of class-discriminative transformer attention. Additionally, \nn+~replaces the GAP method used in the original CAM module by using Global Weighted Ranking Pooling (GWRP), which assigns different weights to the patch tokens based on their relevance scores for each class in the rank.
During training, two classification losses for class scores from class tokens and patch tokens, along with cumulative contrastive losses from all CCT modules, are used to optimize the entire model. 
At inference time, the class-specific transformer attentions (MCT Attention) are combined with PatchCAM maps. The resulting fusion maps are further refined by the PatchAffinity, producing the final class-specific localization maps.
}
\label{fig:overview_c}
\vspace{-8pt}
\end{figure*}

We proposed a novel multi-class token transformer framework to exploit class-specific transformer attention for discriminative object localization for WSSS.
The overall architecture of the proposed \nn, as depicted in \Fig{overview}, includes a transformer encoder and a Class Activation (CAM) module. Unlike standard ViTs with only one class token, the proposed \nn~is equipped with multiple class tokens. This multi-class-token design enables the transformer encoder to learn class-specific localization through the transformer attention between each class token and patch tokens. We also propose to extract the transformer attention between patches to refine the localization maps. To enable the class-specific token learning, we propose a class-aware training strategy to supervise the training of the proposed \nn. The proposed framework can easily integrate the CAM module, which can produce additional class activation maps using the patch tokens to complement the class-specific transformer attention maps. We propose an extended framework, \ie, \nn+ (illustrated in \Fig{overview_c}), which improves \nn~on both the transformer encoder and the CAM module. We introduce the use of the global weighted ranking pooling rather than global average pooling in the conventional CAM module. We also propose a contrastive-class-token module which is inserted into the transformer encoder to enable more discriminative class token learning. Lastly, the final class-specific localization maps are extracted by combining the class-specific transformer attention maps from the transformer encoder and the class activation maps from patch tokens. The following sub-sections will provide a detailed description of these steps.

\subsection{\nn}
\label{sec:3.2}
In this section, we introduce the detailed structure of the proposed multi-class transformer,~\nn~(illustrated in \Fig{overview}), two key uses of its self-attention maps, as well as its variant by integrating the CAM mechanism.

\par\noindent\textbf{Multi-class token structure design.} 
A RGB image is partitioned into $N\times N$ patches. These patches undergo vectorization and a linear projection into a sequence of patch tokens, denoted as $\mathbf{T}_{p} \in \mathbb{R}^{M\times D}$, where $D$ is the embedding dimension and $M=N^2$. We propose the learning of multiple class tokens $\mathbf{T}_{c} \in \mathbb{R}^{C\times D}$, where $C$ denotes the number of classes. These class tokens are then combined with patch tokens and positional embeddings. The resulting tokens $\mathbf{T}_{in} \in \mathbb{R}^{(C+M)\times D}$ are used as input to the transformer encoder, which consists of a series of $L$ encoding layers. Within every transformer encoding layer, a Multi-Head Attention (MHA) module and a Multi-Layer Perceptron (MLP) are utilized. Prior to the MHA and MLP, two LayerNorm layers are applied, one for each, respectively.

\par\noindent\textbf{Class-specific multi-class token attention.} 
In each Multi-Head Attention module, the self-attention mechanism is used to learn the pairwise interactions between tokens. As input, a sequence of tokens is first normalized and then linearly projected to three sequences of vectors representing query $\mathbf{Q} \in \mathbb{R}^{(C+M)\times D}$, key $\mathbf{K} \in \mathbb{R}^{(C+M)\times D}$ and value $\mathbf{V} \in \mathbb{R}^{(C+M)\times D}$. 
The Scaled Dot-Product Attention~\cite{vaswani2017attention} mechanism is used to calculate the attention score between every pair of the query vector and key vector. This attention module updates each token by dynamically aggregating information from all tokens based on their specific attention weights. This is formulated as:
\begin{equation}
    \mathrm{Attention}(\mathbf{Q}, \mathbf{K}, \mathbf{V}) = \mathrm{softmax}(\mathbf{Q}\mathbf{K}^\top/\sqrt{D})\mathbf{V},
\end{equation}
where a global pairwise token attention map $\mathbf{A}_{t2t} \in \mathbb{R}^{(C+M)\times (C+M)} = \mathrm{softmax}(\mathbf{Q}\mathbf{K}^\top/\sqrt{D})$ is obtained, as illustrated by the array of small circles in \Fig{overview}.

As shown in \Fig{overview}, the first $C$ rows of the array of small circles denote the attention scores between each of $C$ classes and all tokens. We extract the yellow part of the array, which denotes the attention scores of each class and all patch tokens, denoted as the class-to-patch attention $\mathbf{A}_{c2p}\in \mathbb{R}^{C\times M}$, where $\mathbf{A}_{c2p} = \mathbf{A}_{t2t}[1:C,C+1:C+M]$. The class-to-patch attention can be transformed into class-specific localization maps by mapping each row of attention scores back to their respective original patch locations.
Each transformer layer has its class-to-patch attention maps. Particularly, deeper (top) layers capture more task-specific high-level representations, while shallow (bottom) layers capture more general low-level representations. In order to strike a balance between the precision and recall of the resulting class-specific localization maps, we propose to combine the class-to-patch attentions from the top $K$ transformer encoding layers.
This process can be formulated as:
\begin{equation}
\setlength{\abovedisplayskip}{3pt}
\setlength{\belowdisplayskip}{3pt}
    \hat{\mathbf{A}}_{mct} = \frac{1}{K}\sum_{l}^{K}\hat{\mathbf{A}}^{l}_{mct},
\end{equation}
where $\hat{\mathbf{A}}^{l}_{mct}$ denotes the class-to-patch attention obtained from the $l^{th}$ layer of the proposed \nn. $\hat{\mathbf{A}}_{mct}$ denotes the resulting combined map. This map is further normalized, generating the final class-specific localization maps $\mathbf{A}_{mct} \in \mathbb{R}^{C\times N\times N}$. The selection of $K$ is extensively discussed~\Sec{experiments} and visualized in~\Fig{lineplot}. 

\begin{figure}[t]
\begin{center}
\includegraphics[width=.48\textwidth]{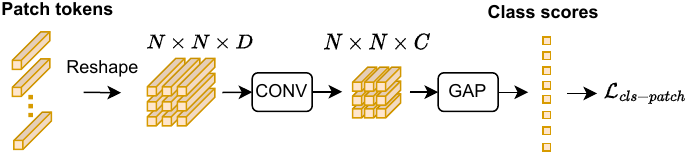}
\end{center}
\vspace{-8pt}
\caption{The classification branch with the CAM module. As input, the patch tokens generated from the last transformer encoding layer are reshaped and then passed through to a convolutional layer (CONV). The resulting feature maps are then processed via Global Average Pooling (GAP) to produce the class scores.}
\label{fig:cam}
\vspace{-8pt}
\end{figure}

\par\noindent\textbf{Class-specific attention refinement.} 
Previous works~\cite{ahn2018learning,wang2020weakly,xu2021leveraging} often utilize pairwise affinity for enhancing object localization maps. Typically, achieving this involves training additional parameters for learning affinities.
In contrast, our approach introduces a novel technique in which a pairwise affinity map is directly obtained from the transformer attention between patches in the proposed \nn, without the need for any additional computations or supervision.
More specifically, we extract the patch-to-patch attentions $\mathbf{A}_{p2p}\in \mathbb{R}^{M\times M}$ from the global pairwise attention map as $\mathbf{A}_{p2p} = \mathbf{A}_{t2t}[C+1:C+M,C+1:C+M]$, which is illustrated by the matrix with blue dots in \Fig{overview}. We then reformat the patch-to-patch attentions into a 4D tensor $\hat{\mathbf{A}}_{p2p}\in \mathbb{R}^{N\times N\times N\times N}$.
This tensor is utilized to enhance the class-specific transformer attention as:
\begin{equation}
\setlength{\abovedisplayskip}{3pt}
    \mathbf{A}_{mct\_ref}(c,i,j) = \sum_{k}^{N}\sum_{l}^{N}\hat{\mathbf{A}}_{p2p}(i,j,k,l)\cdot \mathbf{A}_{mct}(c,k,l),
    \label{attn_refine}
\end{equation}
where $\mathbf{A}_{mct\_ref}\in \mathbb{R}^{C\times N\times N}$ is the refined class-specific localization map. 
As depicted in \Table{abla_mct_pgt} and \Fig{mct_cam}, the utilization of the patch-to-patch attention as the pairwise affinity results in superior class-specific localization maps with enhanced appearance continuity and smoothness. This was not noted in~\cite{gao2021ts}.

\par\noindent\textbf{Class-aware training.} 
Traditional transformers applies a MLP head on the final output class token to predict class scores. Our method involves multiple class tokens denoted as $\mathbf{T}_{cls} \in \mathbb{R}^{C\times D}$. Our goal is to ensure that each class token captures unique and discriminative class-related information. To achieve this, we employ class-wise average pooling on the output class tokens to predict class scores as follows:
\begin{equation}
\setlength{\abovedisplayskip}{2pt}
\setlength{\belowdisplayskip}{1pt}
\mathbf{y}_{cls}^c = \frac{1}{D}\sum_{j}^{D}\mathbf{T}_{cls}(c,j),
\label{eq:cls_token_pred}
\end{equation}
where  $\mathbf{y}_{cls} \in \mathbb{R}^{C}$ denotes the class prediction. 
A multi-label soft margin loss is computed between the class scores and the ground-truth image-level labels as follows:
\begin{equation}
\begin{split}
    &\mathcal{L}_{cls-class}=\text{MultilabelSoftMarginLoss}(\mathbf{y}_{cls},~\mathbf{y}) = \\
    & -\frac{1}{C}\sum_{i=1}^{C}\mathbf{y}^i\log \sigma(\mathbf{y}_{cls}^i)+(1-\mathbf{y}^i)\log(1-\sigma(\mathbf{y}_{cls}^i)).
\end{split}
\setlength\belowdisplayskip{1pt}
\label{MLSM}
\end{equation}
This offers direct class-aware supervision to each class token, enabling them to effectively encapsulate class-specific information.


\par\noindent\textbf{The integration of CAM.} The proposed \nn~can be extended by incorporating a CAM module~\cite{zhang2018adversarial,gao2021ts, zhou2016learning}. We extract the patch tokens $\mathbf{T}_{out\_pat} \in \mathbb{R}^{M\times D}$ from the output tokens of the transformer encoder $\mathbf{T}_{out} \in \mathbb{R}^{(C+M)\times D}$. 
As depicted in \Fig{cam}, the reshaped patch tokens are passed through a convolutional layer with $C$ output channels, generating 2D feature maps denoted as $\mathbf{F}_{out\_pat} \in \mathbb{R}^{N\times N\times C}$. These feature maps are then processed through a Global Average Pooling (GAP) layer to produce class scores. Since the output class tokens also contribute to the class prediction (see Eq.~\eq{cls_token_pred}), the overall loss function comprises two classification losses as follows: 

\begin{equation}
\setlength{\abovedisplayskip}{3pt}
    \mathcal{L}_{total} = \alpha*\mathcal{L}_{cls-class} + \beta*\mathcal{L}_{cls-patch},
\end{equation}
where $\alpha$ and $\beta$ are the loss weights, which are empirically set to 1.

\subsection{\nn+}
\label{sec:3.4}
\Fig{overview_c} presents an overview of the proposed \nn+. The proposed \nn+ aims to enhance both components of \nn, i.e., the transformer encoder and the CAM module, thereby improving their derived class-specific localization maps. This section elaborates on the two enhanced aspects introduced by~\nn+.

\par\noindent\textbf{Global Weighted Ranking Pooling.} CAM~\cite{zhou2016learning} proposed to use Global Average Pooling (GAP) to enable CNNs to have localization ability with only image-level labels. However, Kolesnikov~\etal~\cite{kolesnikov2016seed} revealed that the traditional global pooling techniques have their drawbacks. For instance, Global Average Pooling (GAP) encourages the model to have high responses on all positions, while Global Max Pooling (GMP) encourages the model to have high response on only one position. This thus results in the overestimation and underestimation on the size of object regions by GAP and GMP, respectively. To address these limitations, we introduce the Global Weighted Ranking Pooling (GWRP) method~\cite{kolesnikov2016seed} to the transformer framework to aggregate the patch tokens for class prediction. Unlike GAP, which assigns equal weights to each patch during aggregation, GWRP assigns different weights considering the activation ranking across all patches for each channel:

\begin{equation}
    G_{c}(\textbf{P}) = \frac{1}{S(\lambda)}\sum_{j=1}^{N^{2}}\lambda^{j-1}\textbf{P}^{r_{j},c},
\end{equation}
where $\mathbf{P}\in~\mathbb{R}^{N^2\times C}$ denotes the output by performing channel-wise vectorization on the 2D feature maps derived from the output patch tokens;
$G(\mathbf{F})\in~\mathbb{R}^{C}$ denotes the final pooled result, \ie, the class prediction; $r_{j}$ denotes the ranking index, \eg, $\mathbf{F}^{r_{1},c}>\mathbf{F}^{r_{2},c}>\cdots>\mathbf{F}^{r_{N^2},c}$ for the class $c$; $S(\lambda)= \sum_{j=1}^{N^2}\lambda^{j-1}$, $\lambda$ denotes the decay rate. 
This GWRP strategy allows the model to prioritize the more informative patches, ensuring that they contribute more significantly to the final global class scores. This can mitigate the object overestimation or underestimation issues faced by traditional pooling methods, leading to more accurate and reliable class-specific localization maps.

With the GWRP pooling strategy, the class prediction scores from the output patch tokens can be obtained as:
\begin{align}
    \mathbf{y}_{cls-patch}&=G(\mathbf{T}_{out\_pat}).
\end{align}
This branch ends up with a classification loss between the class prediction and the image-level labels $\mathbf{y}$:
\begin{equation}
    \mathcal{L}_{cls-patch}=\\
    \text{MultilabelSoftMarginLoss}(\mathbf{y}_{patch},~ \mathbf{y}),
\end{equation}
where $\text{MultilabelSoftMarginLoss}$ is given by Eq. (\ref{MLSM}).
\begin{figure}
\begin{center}
\includegraphics[width=.48\textwidth]{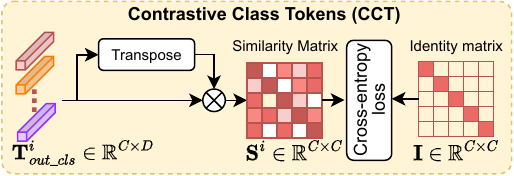}
\end{center}
\vspace{-8pt}
\caption{The Contrastive Class Tokens (CCT) module. The output class tokens $\mathbf{T}^i_{out\_cls}$ of size $C\times D$ from the $i^{th}$ transformer encoding block, are computed with its transposed counterpart via dot product. This generates a similarity matrix with each score representing the pairwise similarity of class tokens. A contrastive loss is imposed on the similarity matrix by computing the cross-entropy loss with an identity matrix as ground-truth, forcing each class token to be different from other tokens.}
\label{fig:cct}
\vspace{-8pt}
\end{figure}
\par\noindent\textbf{Contrastive class token enhancement.} While the class-aware training strategy enables different class tokens to attend to different object regions, the resulting class-to-patch transformer attention maps for different classes present in the same image often contain overlapped localized object regions. This observation can be attributed to two factors: (\textbf{i}) the utilization of the multi-label one-versus-all loss on the class scores, achieved by aggregating each class token embedding, does not strictly ensure discrimination between different class tokens; (\textbf{ii}) By applying the loss only on the highest-level output class tokens, the influence of this loss is primarily limited to the final layers of the network. The middle or lower layers, may not receive sufficient guidance from the loss to effectively differentiate between different class tokens. 

In order to obtain different and non-overlapping class-to-patch transformer attention maps, we propose a Contrastive Class Tokens (CCT) module with a regularization loss that provides stronger supervision and encourages class discrimination on the class tokens. More specifically, as illustrated in Fig.~\ref{fig:cct}, given the output class tokens $\mathbf{T}^i_{out\_cls}\in~\mathbb{R}^{C\times D}$ from the $i^{th}$ transformer encoding layer, we compute the pairwise similarity between every two class tokens, forming a similarity matrix $\mathbf{S}^i\in~\mathbb{R}^{D\times D}$. To encourage each class token to be only similar to itself and dis-similar to all other class tokens, a cross-entropy loss is computed between the similarity matrix and an identity matrix $\mathbf{I}\in~\mathbb{R}^{D\times D}$. This contrastive loss is imposed on the output class tokens of every transformer encoding block. This allows the model to receive stronger and more frequent guidance to learn more discriminative and class-specific representations throughout the network. This contributes to an improved localization performance with the reduced ambiguity arising from overlapping attention regions. The similarity matrix and the proposed contrastive loss are defined as follows:
\begin{align}
    \mathbf{S}^{i} &= \mathbf{T}^i_{out\_cls}\cdot(\mathbf{T}^i_{out\_cls})^\intercal, \\
    \mathcal{L}_{reg} &= \frac{1}{L}\sum_{i=1}^{L}CrossEntropyLoss(\mathbf{S}^i, \mathbf{I}),
\end{align}
where $i$ is the index of the transformer encoding layer and $L$ is the number of the transformer encoding layers.

The total loss is formulated by a combination of two classification losses and one regularization loss as follows:
\begin{equation}
\setlength{\abovedisplayskip}{3pt}
    \mathcal{L}_{total} = \alpha*\mathcal{L}_{cls-class} + \beta*\mathcal{L}_{cls-patch} + \gamma*\mathcal{L}_{reg}.
\end{equation}
where $\alpha$, $\beta$ and $\gamma$ are the loss weights, which are empirically set to 1.

\subsection{Class-specific localization inference}
\par\noindent\textbf{Map fusion.} 
At the inference stage, the CAM maps derived from patch tokens (denoted as PatchCAM, $\mathbf{A}_{pCAM}\in \mathbb{R}^{N\times N\times C}$) can be obtained the convolutional layer of the CAM module. This is achieved by normalizing the feature map $\mathbf{F}_{out\_pat}$. 
The PatchCAM maps can act as a complement to the proposed multi-class transformer attention maps. The improved class-specific localization maps $\mathbf{A}$ can be obtained by a combination of these two map types: 
\begin{equation}
\setlength{\abovedisplayskip}{3pt}
\setlength{\belowdisplayskip}{3pt}
    \mathbf{A} = \mathbf{A}_{pCAM} \circ \mathbf{A}_{mct},
\end{equation}
where $\circ$ denotes the element-wise multiplication operator. 

\par\noindent\textbf{Map refinement.}
The fused class-specific localization maps can be further refined by the proposed affinities, \ie, the transformer attention between patch tokens: 
\begin{equation}
\setlength{\abovedisplayskip}{3pt}
\setlength{\belowdisplayskip}{3pt}
    \mathbf{A}_{ref}(c,i,j) = \sum_{k}^{N}\sum_{l}^{N}\hat{\mathbf{A}}_{p2p}(i,j,k,l)\cdot \mathbf{A}(c,k,l).
\end{equation}
The proposed \nn+~introduces an effective transformer-based framework that allows the CAM module to adapt flexibly and robustly to multi-label images.
By applying the classification loss on class predictions from both class tokens and patch tokens, the proposed framework enforces a strong consistency between these two types of tokens, thereby enhancing the model learning. The intuition is mainly two-fold. First, this consistency constraint acts as an auxiliary supervision, guiding the learning process towards more effective patch representations. Second, the strong pairwise interaction, achieved through message passing, between the patch tokens and the multiple class tokens leads to to more representative patch tokens. Consequently, this generates more class-discriminative PatchCAM maps, surpassing the results obtained by solely using one class token as seen in TS-CAM~\cite{gao2021ts}.

\section{Experiments}
\label{sec:experiments}
\subsection{Experimental Settings}
\par\noindent\textbf{Datasets.} Three datasets including PASCAL VOC 2012~\cite{everingham2010pascal}, MS COCO 2014~\cite{lin2014microsoft} and OpenImages~\cite{choe2020evaluating} were used to evaluate the proposed method. \textbf{PASCAL VOC} is divided into training (\textit{train}) ($1,464$ images),  validation (\textit{val}) ($1,449$ images) and test ($1,456$ images) sets, respectively. There are 21 categories including one background class in this dataset. A widely adopted training practice~\cite{chang2020weakly,wang2020self,lee2021anti,su2021context,zhang2021complementary,xu2021leveraging} involves incorporating extra data from~\cite{hariharan2011semantic} to compose a total number of 10,582 training images.
\textbf{MS COCO} is divided into training (80K images) and validation (40K images) sets. There are 81 categories including on background class.
\textbf{OpenImages} is divided into training (29,819 images), validation (2,500 images) and test (5,000 images) sets. There are 100 categories in this dataset.  


\begin{table}[t]
\caption{Comparison of mIoUs (\%) for the seed maps and pseudo semantic masks on the \textit{train} sets. }
\vspace{-8pt}
\label{tab:seed}
\small
\centering
\resizebox{\linewidth}{!}
{\begin{tabular}{lcccc}
\toprule 
\multirow{2}{*}{Method} & \multicolumn{2}{c}{VOC} & \multicolumn{2}{c}{COCO} \\
& Seed       & Mask       & Seed        & Mask       \\
\midrule
PSA (CVPR18) \cite{kolesnikov2016seed} & 48.0 & 61.0 &-&-\\
IRN (CVPR19) \cite{ahn2019weakly} & 48.8&66.5&33.1& 42.5 \\
Chang \etal (CVPR20) \cite{chang2020weakly} & 50.9&63.4&-&- \\
SEAM (CVPR20) \cite{wang2020self} &55.4& 63.6&25.1&31.5\\
CONTA (NeurIPS)~\cite{zhang2020causal} & 48.8 &67.9 & 28.7&35.2 \\
CDA (ICCV21) \cite{su2021context} &58.4&66.4&-&- \\
Zhang \etal (ICCV21) \cite{zhang2021complementary} &57.4& 67.8&-&- \\
RIB (NeurIPS21)~\cite{lee2021reducing}&56.5&70.6&36.5& 45.6\\
CLIMS (CVPR22)~\cite{xie2022clims}&56.6&70.5&-&- \\
SIPE (CVPR22)~\cite{chen2022self}&58.6&64.7&-&-\\
W-OoD (CVPR22)~\cite{lee2022weakly}&59.1&72.1&-&-\\
Du~\textit{et al.} (CVPR22)~\cite{du2022weakly}&61.5&70.1&-&-\\
AMN (CVPR22)~\cite{lee2022threshold}& 62.1&72.2&40.3&46.7 \\
AdvCAM (PAMI22) \cite{lee2022anti} &55.6& 69.9&37.2&46.0\\
Yoon~\etal~(ECCV22)~\cite{yoon2022adversarial} & 56.0&71.0&-&- \\
LPCAM (CVPR23)~\cite{chen2023extracting} & 65.3 & 72.7 & 42.5 & 47.7 \\

 \midrule
MCTformer~\cite{xu2022multi}& 61.7 & 69.1&36.6&41.6 \\
\textbf{MCTformer+} & \textbf{68.8} & \textbf{76.2}& \textbf{42.8}&\textbf{48.1} \\
  \bottomrule 
\end{tabular}
}
\vspace{-8pt}
\end{table}
\begin{table}[t]
\caption{Comparison of mIoUs(\%) for segmentation performance on MS COCO \textit{val} set. Network denotes the network for semantic segmentation. Sup. denotes weak supervision. I denotes image-level labels; S denotes saliency maps.}
\vspace{-8pt}
\label{tab:coco}
\small
\centering
\begin{tabular}{lccc}
\toprule 
Method            & Network  &Sup.          & Val                       \\ \midrule
EPS (CVPR21) \cite{lee2021railroad} & V2-VGG16 &I+S & 35.7 \\
RCA (CVPR22)~\cite{zhou2022regional} & V2-VGG16 & I+S & 36.8 \\
AuxSegNet (ICCV21) \cite{xu2021leveraging} & V1-RN38 & I+S & 33.9 \\ 
L2G (CVPR22)~\cite{jiang2022l2g} & V2-RN101 & \textbf{I+S} & \textbf{44.2} \\
  \midrule
Wang \etal (IJCV20) \cite{wang2020weakly}&V2-VGG16&I&27.7\\
    Luo \etal (AAAI20)  \cite{luo2020learning}  &     V2-VGG16              &  I&    29.9 \\
SEAM (CVPR20) \cite{wang2020self} & V1-RN38 & I&31.9 \\
CONTA (NeurIPS20) \cite{zhang2020causal} & V1-RN38 & I&32.8 \\
Kweon \etal (ICCV21) \cite{kweon2021unlocking} & V1-RN38 & I & 36.4 \\
CDA (ICCV21) \cite{su2021context} & V1-RN38 & I &33.2 \\
AdvCAM (CVPR21)~\cite{lee2021anti} & V2-RN101 & I & 44.4 \\
Li~\etal~(CVPR22)~\cite{li2022towards}& V2-RN101& I & 44.7 \\
AMN (CVPR22)~\cite{lee2022threshold} & V2-RN101 & I & 44.7 \\
SIPE (CVPR22)~\cite{chen2022self} & V1-RN38 &I&43.6 \\
Yoon~\etal (ECCV22)~\cite{yoon2022adversarial} &V1-RN38 & I & 44.8\\
OCR (CVPR23)~\cite{cheng2023out} & V1-RN38&I&42.5 \\
 BECO (CVPR23)~\cite{rong2023boundary} & V3+-RN101 & I&45.1 \\
LPCAM (CVPR23)~\cite{chen2023extracting} & V1-RN38 & I & 42.8 \\
 \midrule
 \nn~\cite{xu2022multi} & V1-RN38 &I & 42.0 \\
  \textbf{\nn+} & V1-RN38 &\textbf{I}& \textbf{45.2} \\
  \bottomrule 
\end{tabular}
\label{tab:segsota}
\vspace{-8pt}
\end{table}
\begin{table}[t]
\caption{Comparisons of mIoUs (\%) for segmentation performance on PASCAL VOC. 
\label{tab:sota_res38}}
\vspace{-8pt}
\centering
\small
\resizebox{1.0\linewidth}{!}{
\begin{tabular}{lcccc}
\toprule
Method            & Network  & Sup.          & Val            & Test           \\ \midrule
  CIAN (AAAI20)  \cite{fan2020cian}    &         V2-RN101     &     I+S&  64.3         &  65.3 \\
ICD (CVPR20) \cite{fan2020learning}    &          V2-RN101          &   I+S&    67.8         &    68.0           \\ 
 Zhang \etal (ECCV20) \cite{zhang2020splitting}   &       ResNet50  &   I+S&   66.6         &   66.7    \\  
 Sun \etal (ECCV20) \cite{sun2020mining}   &          V2-RN101        &I+S&        66.2          &      66.9        \\
 EDAM (CVPR21) \cite{wu2021embedded} & V2-RN101 & I+S & 70.9 & 70.6 \\
 EPS (CVPR21) \cite{lee2021railroad} & V2-RN101 & I+S & 71.0&71.8  \\
 Yao \etal (CVPR21) \cite{yao2021non} & V2-RN101 & I+S &68.3&68.5 \\
 AuxSegNet (ICCV21) \cite{xu2021leveraging} & V1-RN38 & I+S &69.0&68.6 \\  
 Li~\etal~(CVPR22)~\cite{li2022towards}&V2-RN101&I+S&72.0&72.9 \\
 Du~\etal~(CVPR22)~\cite{du2022weakly}&V2-RN101&\textbf{I+S}&\textbf{72.6}&\textbf{73.6} \\
 RCA (CVPR22)~\cite{zhou2022regional} & V2-RN38&I+S&72.2&72.8 \\
 L2G (CVPR22)~\cite{jiang2022l2g} & V1-RN38&I+S&72.0&73.0\\
  \midrule
  Zhang \etal (AAAI20) \cite{zhang2019reliability} & V1-RN38 & I&62.6 & 62.9 \\
    Luo \etal (AAAI20)  \cite{luo2020learning}  &     V2-RN101         &I     &       64.5       & 64.6 \\
Chang \etal (CVPR20) \cite{chang2020weakly}  &            V2-RN101    &I     &      66.1        &    65.9\\ 
Araslanov \etal (CVPR20) \cite{araslanov2020single}    &          V1-RN38    &I     &      62.7        &   64.3       \\ 
SEAM (CVPR20) \cite{wang2020self} & V1-RN38 &I& 64.5 & 65.7 \\
 BES (ECCV20) \cite{chen2020weakly} & V2-RN101 & I &65.7&66.6 \\
 CONTA (NeurIPS20) \cite{zhang2020causal}   &          V1-RN38         &  I&    66.1       &    66.7            \\  
AdvCAM (CVPR21) \cite{lee2021anti} & V2-RN101 & I &68.1 &68.0 \\
 ECS-Net (ICCV21) \cite{sun2021ecs} & V1-RN38 & I &66.6 & 67.6 \\
 Kweon \etal (ICCV21) \cite{kweon2021unlocking} & V1-RN38 & I &68.4&68.2 \\
 CDA (ICCV21) \cite{su2021context} & V1-RN38 & I &66.1&66.8 \\
 Zhang \etal (ICCV21) \cite{zhang2021complementary} & V1-RN38 & I & 67.8&68.5 \\
 AMN (CVPR22)~\cite{lee2022threshold} & V2-RN101 & I & 70.7 & 70.6 \\
 W-OoD (CVPR22)~\cite{lee2022weakly}& V1-RN38 & I & 70.7& 70.1 \\
 SIPE (CVPR22)~\cite{chen2022self} & V1-RN38 & I & 68.2 & 69.7 \\
 Yoon~\textit{et al.} (ECCV22)~\cite{yoon2022adversarial} & V1-RN38 & I &70.9 & 71.7 \\
 OCR (CVPR23)~\cite{cheng2023out} & V1-RN38 & I & 72.7 & 72.0 \\
 BECO (CVPR23)~\cite{rong2023boundary} & V3+-RN101 & I & 72.1 & 71.8 \\
 LPCAM (CVPR23)~\cite{chen2023extracting} & V1-RN38 & I & 72.6 & 72.4 \\
 ToCo (CVPR23)~\cite{ru2023token} & ViT-B & I & 71.1 & 72.2 \\
 
 \midrule
  \nn~\cite{xu2022multi}  & V1-RN38 &I&   71.9 &  71.6\\
    \textbf{\nn+}  & V1-RN38 &\textbf{I}&    \textbf{74.0} &  \textbf{73.6}\\
  \bottomrule
\end{tabular}
}
\vspace{-8pt}
\end{table}

\par\noindent\textbf{Evaluation metrics.}
For the WSSS task, following prior studies~\cite{lee2021anti}, the mean Intersection-over-Union (mIoU) was used to evaluate both the quality of the generated pseudo semantic masks on the \textit{train} sets and the final semantic segmentation performance on the \textit{val} sets of both PASCAL VOC and MS COCO datasets. The official online evaluation server was used to produce the segmentation results on PASCAL VOC \textit{test} set.
For the WSOL task, our focus was on the pixel-level evaluation. By following previous studies~\cite{choe2020evaluating,zhu2022weakly}, the peak Intersection-over-Union (pIoU) and the pixel average precision (PxAP) were used to evaluate the class-specific localization maps on the \textit{test} set.

\par\noindent\textbf{Implementation details.}
To construct the proposed multi-class token transformer, we used DeiT-S as the backbone~\cite{touvron2021training, gao2021ts}.
We used the pre-trained weights of DeiT-S on ImageNet~\cite{deng2009imagenet} for model initialization. In particular, all class tokens were initialized by the pre-trained class token weights of DeiT-S. We used the data augmentation methods and other training hyper-parameters suggested in~\cite{touvron2021training, gao2021ts}. 
The decay parameter in GWRP was set to 0.996 as suggested in~\cite{kolesnikov2016seed}.
For the semantic segmentation network, we followed previous studies~\cite{ahn2018learning, zhang2019reliability, xu2021leveraging, zhang2021complementary} to use DeepLab-V1 with ResNet38~\cite{wu2019wider} as the backbone. At the inference stage, post-processing with multi-scale inputs and CRFs~\cite{chen2014semantic} was used.

\begin{figure*}
\begin{center}
\includegraphics[width=.88\textwidth]{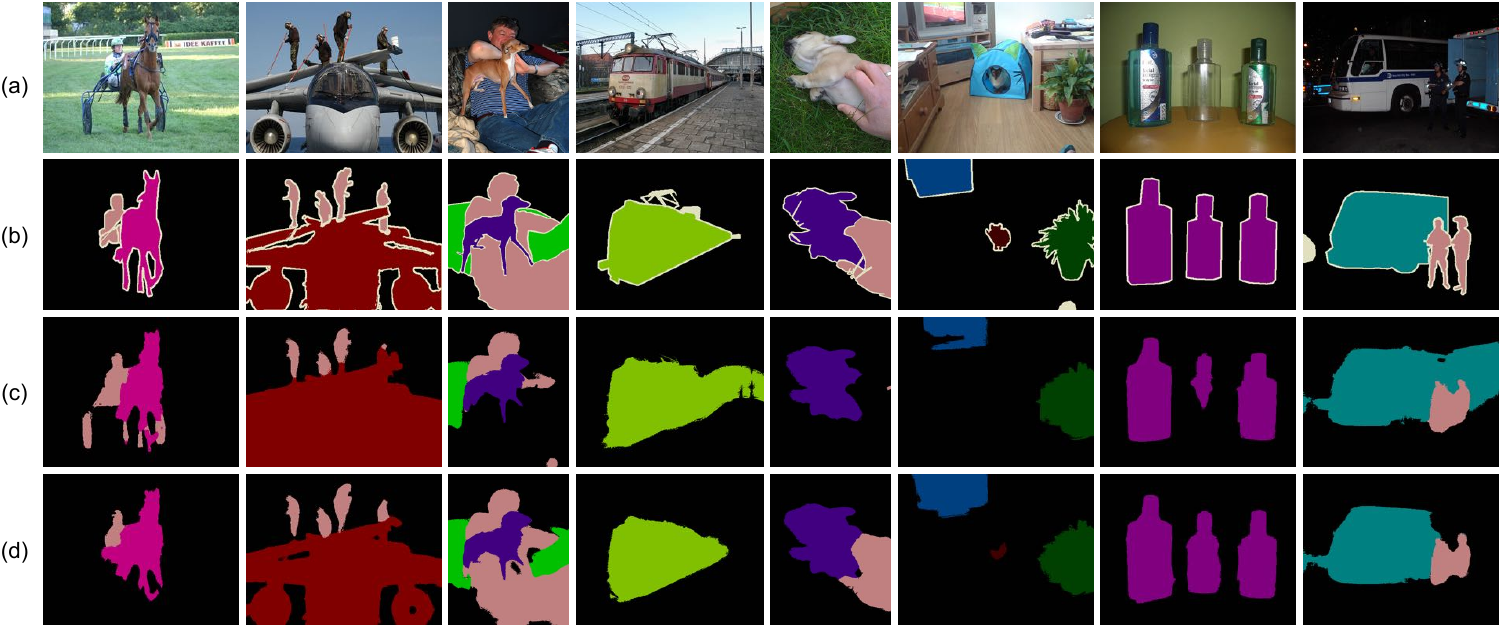}
\vspace{-8pt}
\end{center}
\caption{Visualization of segmentation results on PASCAL VOC \textit{val} set. (a) Input RGB images. (b) Ground-truth segmentation masks. (c) Segmentation results of MCTformer~\cite{xu2022multi}. (d) Segmentation results of MCTformer+.}
\label{fig:vocsegresults}
\vspace{-8pt}
\end{figure*}
\begin{figure*}
\begin{center}
\includegraphics[width=.88\textwidth]{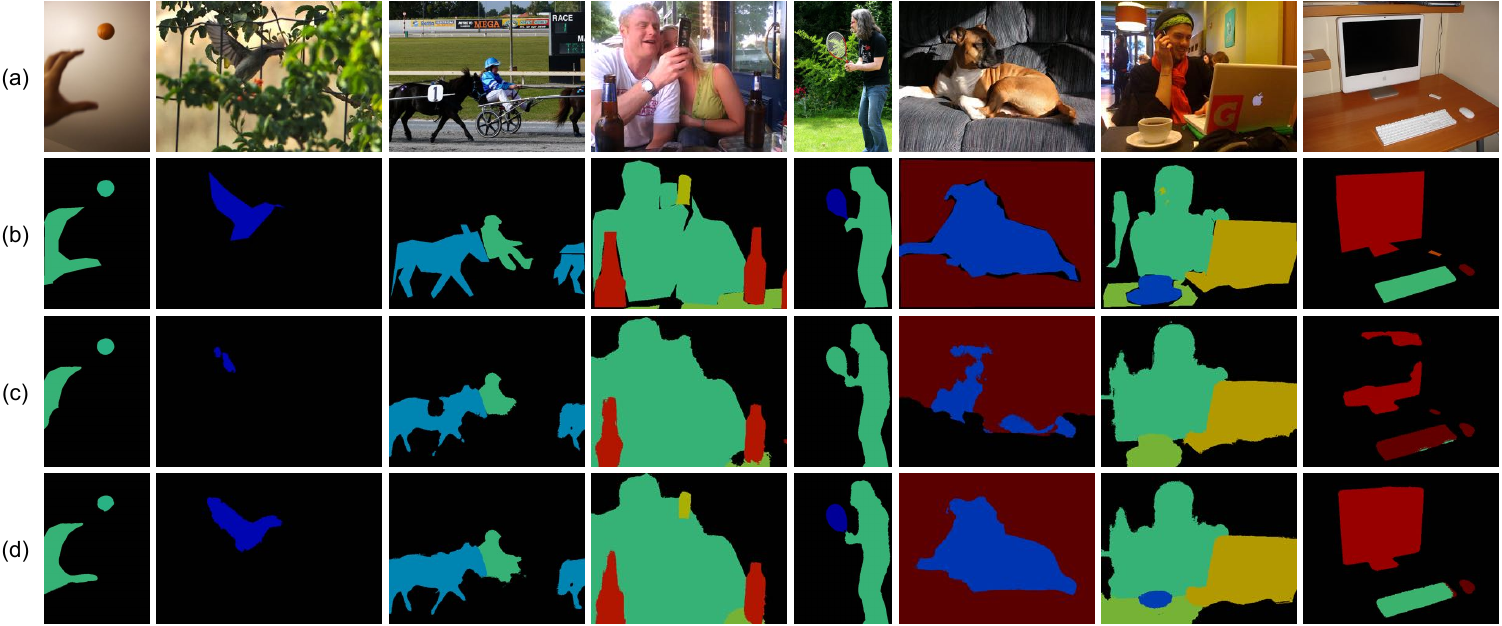}
\vspace{-8pt}
\end{center}
\caption{Visualization segmentation results on MS COCO \textit{val} set. (a) Input RGB images. (b) Ground-truth segmentation masks. (c) Segmentation results of MCTformer~\cite{xu2022multi}. (d) Segmentation results of MCTformer+.}
\label{fig:cocosegresults}
\vspace{-8pt}
\end{figure*}
\begin{table*}[t]
\caption{Per-class segmentation IoUs (\%) on PASCAL VOC. $^*$ denotes no post-processing.}
\vspace{-8pt}
\label{tab:perclass_iou_voc}
\small
\centering
\resizebox{\linewidth}{!}
{\begin{tabular}{lcccccccccccccccccccccc}
\toprule 
&bkg&plane&bike&bird&boat&bottle&bus&car&cat&chair&cow&table&dog&horse&mbk&person&plant&sheep&sofa&train&tv&mIoU \\
\midrule
 \multicolumn{23}{l}{Results on the \textit{val} set:} \\
\nn$^*$~\cite{xu2022multi}& 90.6&71.8&37.5&85.1&52.9&68.8&78.8&78.7&87.1&28.4&78.9&53.0&83.9&78.2&76.8&76.4&54.1&80.1&46.0&71.6&54.3&68.2\\
\textbf{\nn+}$^*$&92.1&81.0&36.6&86.3&62.0&75.0&85.5&82.0&87.8&29.0&79.0&56.6&83.5&77.2&76.4&80.0&52.1&81.2&48.9&81.4&51.9&70.7\\
\nn~\cite{xu2022multi}&91.9&78.3&\textbf{39.5}&89.9&55.9&76.7&81.8&79.0&90.7&\textbf{32.6}&\textbf{87.1}&57.2&\textbf{87.0}&\textbf{84.6}&77.4&79.2&\textbf{55.1}&\textbf{89.2}&47.2&70.4&\textbf{58.8}&71.9 \\
\textbf{\nn+} &\textbf{93.3}&\textbf{87.0}&37.8&\textbf{91.1}&\textbf{66.8}&\textbf{79.9}&\textbf{87.4}&\textbf{82.2}&\textbf{91.3}&32.1&84.8&\textbf{58.8}&86.2&82.2&\textbf{79.0}&\textbf{82.2}&54.4&87.5&\textbf{50.0}&\textbf{82.0}&57.3&\textbf{74.0}\\
 \midrule
  \multicolumn{23}{l}{Results on the \textit{test} set:} \\
\nn$^*$~\cite{xu2022multi}&90.9&76.0&37.2&79.1&54.1&69.0&78.1&78.0&86.1&30.3&79.5&58.3&81.7&81.1&77.0&76.4&49.2&80.0&55.1&65.4&54.5&68.4 \\
\textbf{\nn+}$^*$&92.1&84.0&36.2&79.0&58.5&68.5&84.3&80.9&86.7&32.0&80.7&60.0&82.5&83.2&80.0&78.8&54.1&83.2&55.6&77.4&53.1&71.0 \\
\nn~\cite{xu2022multi}&92.3&84.4&37.2&82.8&60.0&72.8&78.0&79.0&89.4&31.7&\textbf{84.5}&59.1&85.3&83.8&79.2&\textbf{81.0}&53.9&85.3&\textbf{60.5}&65.7&57.7&71.6\\
\textbf{\nn+}&\textbf{93.1}&\textbf{88.8}&\textbf{37.7}&\textbf{82.9}&\textbf{61.4}&\textbf{74.2}&\textbf{85.3}&\textbf{81.0}&\textbf{89.6}&\textbf{33.2}&83.8&\textbf{60.4}&\textbf{85.7}&\textbf{84.4}&\textbf{82.9}&80.8&\textbf{58.9}&\textbf{85.7}&59.1&\textbf{79.4}&\textbf{57.8}&\textbf{73.6} \\
  \bottomrule 
\end{tabular}
}
\end{table*}

\subsection{Comparison with WSSS state-of-the-art}
\label{sec:sota-wsss}
\par\noindent\textbf{PASCAL VOC.} 
We evaluated the mIoUs of the class-specific localization maps (denoted as seed) generated by the proposed method and their post-processed outputs (denoted as mask) by using the refinement methods~\cite{ahn2018learning, ahn2019weakly} following a common practice~\cite{chang2020weakly,wang2020self,lee2021anti,su2021context,zhang2021complementary}, where the masks are the pseudo labels for training segmentation networks.
As depicted in~\Table{seed}, the proposed~\nn+ demonstrates substantial performance improvements over existing methods in terms of both seeds and masks. Particularly, the proposed method outperforms the best method (LPCAM)~\cite{chen2023extracting} in seed and MCTformer~\cite{xu2022multi} by 3.5\% and 7.1\%, respectively. The segmentation mIoUs achieved by the proposed \nn+ are 74.0\% and 73.6\% on the \textit{val} and \textit{test} sets, as shown in \Table{sota_res38}. 
Remarkably, the proposed \nn+ method surpasses all existing methods that solely utilize image-level labels, even achieving comparable or superior results to the methods that use additional saliency maps.
More detailed per-class IoUs can be found in \Table{perclass_iou_voc}, which shows that in both the \textit{val} and \textit{test} sets, the proposed MCTformer+ outperforms MCTformer in most classes, when either using or not using post-processing methods. \Fig{vocsegresults} presents the qualitative segmentation comparison between MCTformer+ and MCTformer on the \textit{val} set. 
It shows that MCTformer+ yields a more discriminative segmentation model, capable of effectively identifying objects even in challenging scenarios such as occlusion. For instance, it successfully distinguishes the sofa in the third column, overcoming occlusion to accurately locate the object of interest. Moreover, MCTformer+ can well segment transparent objects, as observed in the seventh column where it accurately discriminates the bottle in the middle. Furthermore, the model is also able to segment small objects, exemplified by its successful identification of the cat in the sixth column. These results highlight the robustness and the efficacy of the proposed MCTformer+.

\par\noindent\textbf{MS COCO.} 
As reported in \Table{coco}, the proposed \nn+ obtained a segmentation mIoU of 45.2\%, achieving a superior result than the recent methods. 
A notable observation is that a few methods employing extra saliency information yield lower performance than recent methods that solely rely on image-level labels. This highlights the limitations of using pre-trained saliency models, which might exhibit sub-optimal performance on complex datasets. We provide qualitative segmentation results in \Fig{cocosegresults}.

\begin{table}[t]
\caption{Comparison of model complexity.}
\vspace{-8pt}
\label{tab:complexity}
\small
\centering
\begin{tabular}{lcccc}
\toprule 
Model&Image size&\#Params (M)&MACs (G) \\ \midrule
ResNet38&$224\times224$&104.3&99.8 \\
\nn &$224\times224$&21.8&4.7 \\
  \bottomrule 
\end{tabular}
\vspace{-4pt}
\end{table}
\begin{table}[t]
\caption{Comparison of pIoUs (\%) and PxAPs (\%) for the pixel-wise localization maps on OpenImages test set. }
\vspace{-1em}
\label{openimages}
\centering
\resizebox{0.95\linewidth}{!}{
\begin{tabular}{lccc}
\toprule 
Method     &Cls. backbone  &    pIoU  & PxAP                   \\ \midrule
CAM (CVPR16)~\cite{zhou2016learning} &ResNet50&43.0 &58.2 \\
HAS (ICCV17)~\cite{singh2017hide}&ResNet50&41.9 & 55.1 \\
ACoL (CVPR18)~\cite{zhang2018adversarial} &ResNet50& 41.7 & 56.4 \\
SPG (ECCV18)~\cite{zhang2019self} &ResNet50&41.8&55.8\\
ADL (CVPR19)~\cite{choe2019attention} &ResNet50&42.1&55.0\\
CutMix (ICCV19)~\cite{yun2019cutmix} &ResNet50&42.7&57.6\\
PAS (ECCV20)~\cite{bae2020rethinking} &ResNet50&-&60.9\\
IVR (ICCV21)~\cite{kim2021normalization} &ResNet50&-&58.9\\
Zhu~\textit{et al.} (CVPR22)~\cite{zhu2022weakly} &ResNet50&49.7&65.4\\
CREAM (CVPR22)~\cite{xu2022cream} &ResNet50&-&64.7\\
Zhu~\textit{et al.} (ECCV22)~\cite{zhu2022bagging}&ResNet50&52.2&67.7 \\
 \midrule
\textbf{MCTformer+}&ViT-small&\textbf{55.6}&\textbf{72.8} \\
  \bottomrule 
\end{tabular}}
\vspace{-4pt}
\end{table}

\par\noindent\textbf{Model complexity.} \Table{complexity} provides a comparison of the model complexity between the proposed MCTformer based on DeiT-S and a widely used model, ResNet38~\cite{wu2019wider}, for producing class-specific localization maps~\cite{ahn2018learning,wang2020self,zhang2021complementary}. The comparison is based on the number of parameters and multiply-add calculations (MACs). \Table{complexity} demonstrates that the proposed \nn~exhibits significantly lower complexity than ResNet38.


\subsection{Comparison with WSOL State-of-the-art} 
\label{sec:sota-wsol}
\par\noindent\textbf{OpenImages.} We evaluated dense localization performance of the proposed \nn+ on a recently introduced challenging multi-class single-label OpenImages dataset with complex background context. Table~\ref{openimages} shows that our method achieved a pIoU of 55.6\% and a PxAP of 72.8\%, outperforming previous methods by significant margins.

\subsection{Ablation Studies on \nn}
\label{sec:abla}
\begin{figure*}
\begin{center}
\includegraphics[width=.95\textwidth]{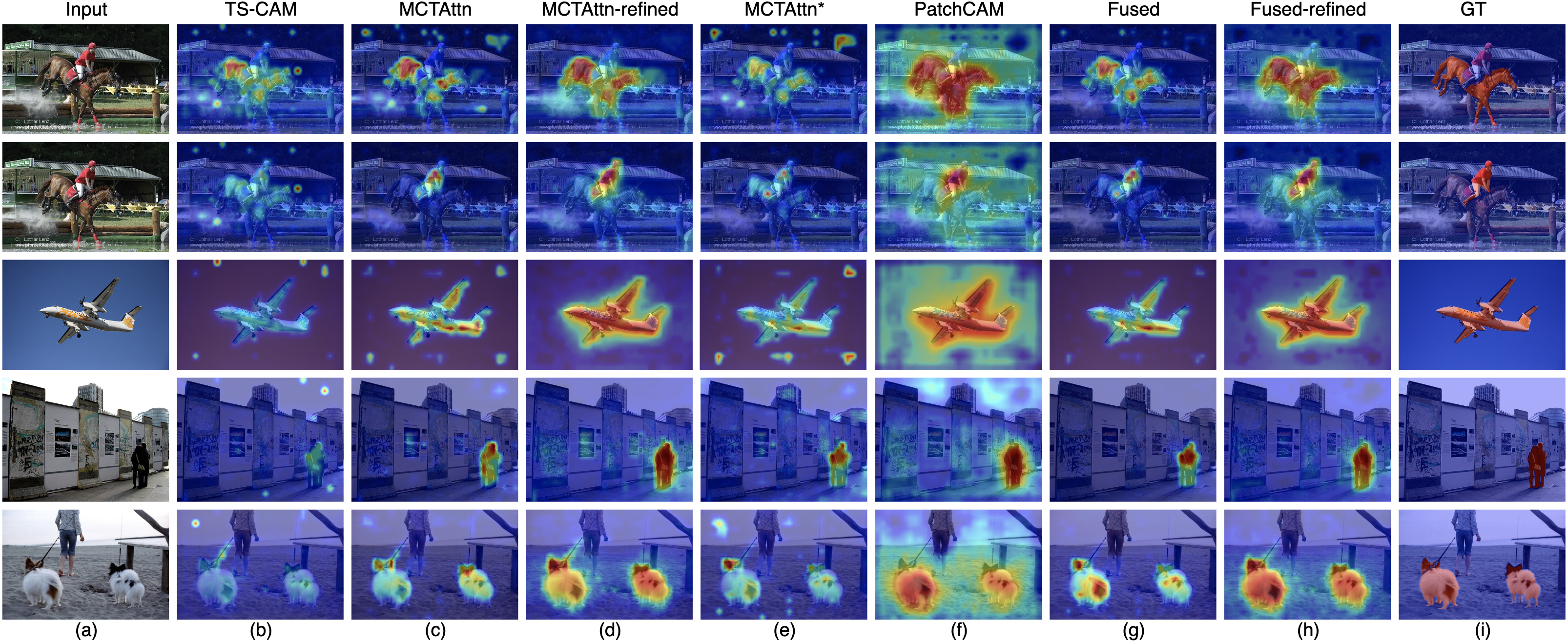}
\vspace{-8pt}
\end{center}
\caption{Comparison of the qualitative localization results by different methods. (b) TS-CAM~\cite{gao2021ts}; (c) \textbf{MCTAttn}: class-specific transformer  attention maps by the proposed \nn~\cite{xu2022multi}; (d) \textbf{Attn-refined}: results of refining MCTAttn using the proposed patch affinity; (e) \textbf{MCTAttn*}: MCTAttn by the extended version of \nn~with the CAM module; (f) \textbf{PatchCAM}: the PatchCAM maps by \nn~with CAM. (g) \textbf{Fused}: results of fusing MCTAttn* and PatchCAM; (h) \textbf{Fused-refined}: the refined fusion maps using the patch affinity from \nn~with CAM. (i) Ground-truth.}
\label{fig:mct_cam}
\vspace{-8pt}
\end{figure*}

\par\noindent\textbf{Effect of multi-class token learning}.
Traditional ViTs with one class token can only produce localization maps independent of classes. TS-CAM~\cite{gao2021ts} achieved class-specific localization maps by applying CAM on patch tokens of a ViT. Without any modification on their official implementation, we only obtained 29.9\% in mIoU by using TS-CAM. By simply adding a ReLU layer on their CAM maps, TS-CAM (denoted as TS-CAM$^\ast$) resulted in a substantial gain of 11.4\%, as reported in \Table{abla_mct_pgt}. 
In contrast, our proposed baseline network, \nn, which generates class-specific localization maps from the transformer attention between each of multiple class tokens and patch tokens, achieved 47.2\% in mIoU, surpassing TS-CAM$^\ast$ by a notable margin of 5.9\%. This result indicates the advantage of the class-specific transformer attention by learning multiple class tokens.


\par\noindent\textbf{Effect of map fusion}. 
\Table{abla_mct_pgt} shows that the fusion of the class-specific transformer attention maps and the CAM maps derived from the patch tokens in the extended version of the proposed \nn, leads to a localization mIoU of 58.2\%. By further refining the fused localization maps by utilizing the patch-to-patch attention as affinity, the resulting mIoU improved significantly, reaching 61.7\%.
In \Fig{mct_cam}, it is evident that the class-specific transformer attention effectively localizes objects, although it exhibits lower responses and some noise (e). Conversely, the PatchCAM maps (f) show higher responses in object regions, but they also activate more background pixels around the objects. Fusing these two map types resulted in significantly improved localization maps (g), which concentrate solely on object regions while substantially reducing background noise. This demonstrates the remarkable superiority of our proposed method over TS-CAM~\cite{gao2021ts} (b), which exhibits sparse and low object responses in most cases.

\begin{table}[t]
\caption{Comparison of mIoUs(\%) for the seed maps generated by different methods on PASCAL VOC \textit{train} set. }
\vspace{-8pt}
\label{tab:abla_mct_pgt}
\small
\centering
\resizebox{1.0\linewidth}{!}{
\begin{tabular}{lc}
\toprule 
Method           & mIoU                   \\ \midrule
TS-CAM~\cite{gao2021ts} &29.9\\
TS-CAM$^\ast$~\cite{gao2021ts} & 41.3 \\
\nn~(Attention) & 47.2 \\
\nn~(Attention + PatchAffinity) & 55.2 \\
\nn~(Attention + PatchCAM) &58.2 \\
\nn~(Attention + PatchCAM + PatchAffinity) & \textbf{61.7} \\ 
\bottomrule
\end{tabular}
}
\vspace{-4pt}
\end{table}

\begin{table}[t]
\caption{Comparison of mIoUs(\%) for segmentation results led by different methods on PASCAL VOC \textit{val} set. }
\vspace{-8pt}
\label{tab:abla_mct_seg}
\small
\centering
\resizebox{1.0\linewidth}{!}{
\begin{tabular}{lc}
\toprule 
Method           & mIoU                   \\ \midrule
TS-CAM$^\ast$~\cite{gao2021ts} & 49.7 \\
\nn~(Attention) & 55.6 \\
\nn~(Attention + PatchAffinity) & 58.8 \\
\nn~(Attention + PatchCAM) &61.1 \\
\nn~(Attention + PatchCAM + PatchAffinity) & \textbf{62.6} \\ 
\bottomrule
\end{tabular}
}
\vspace{-4pt}
\end{table}
\begin{table}[t]
\caption{Comparison of mIoUs(\%) for the class-specific transformer attention by different methods of class prediction in the proposed~\nn~on PASCAL VOC \textit{train} set. }
\vspace{-8pt}
\label{tab:average-pool}
\small
\centering
\begin{tabular}{lccc}
\toprule 
          & Fully-connected  & Max-pooling & Average-pooling                   \\ \midrule
mIoU &41.5&26.8&\textbf{47.2}\\
  \bottomrule 
\end{tabular}
\end{table}
\begin{figure}
\begin{center}
\includegraphics[width=.4\textwidth]{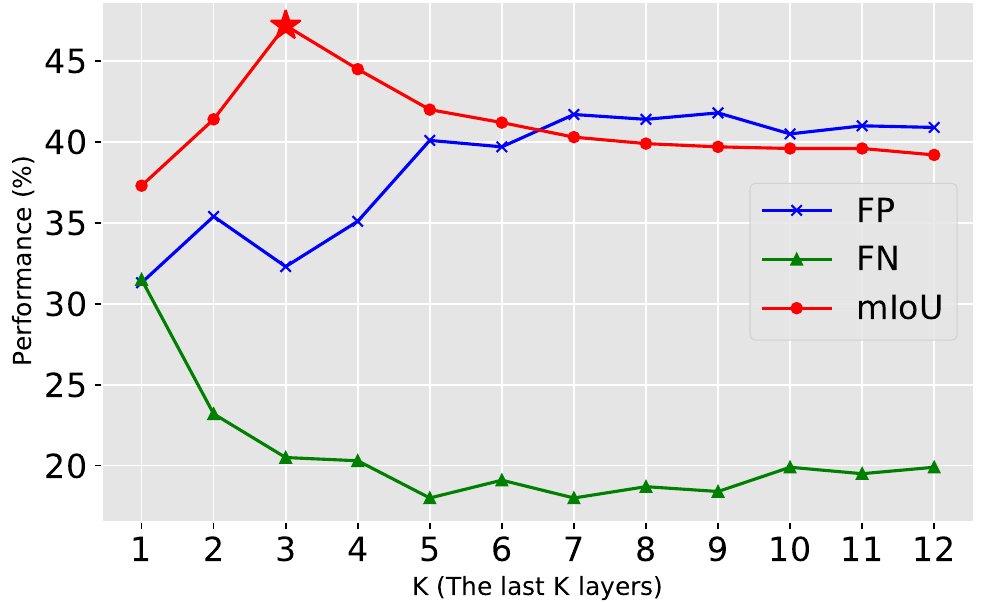}
\end{center}
\vspace{-8pt}
\caption{Comparison of the false positive rates (FP)(\%), false negative rate (FN)(\%) and mIoU(\%) for the class-specific transformer attention maps by the proposed \nn. The x-axis $K$ represents the number of layers (from top to bottom) from which the attention maps are fused.}
\label{fig:lineplot}
\vspace{-8pt}
\end{figure}

\begin{figure*}
\begin{center}
\includegraphics[width=.88\textwidth]{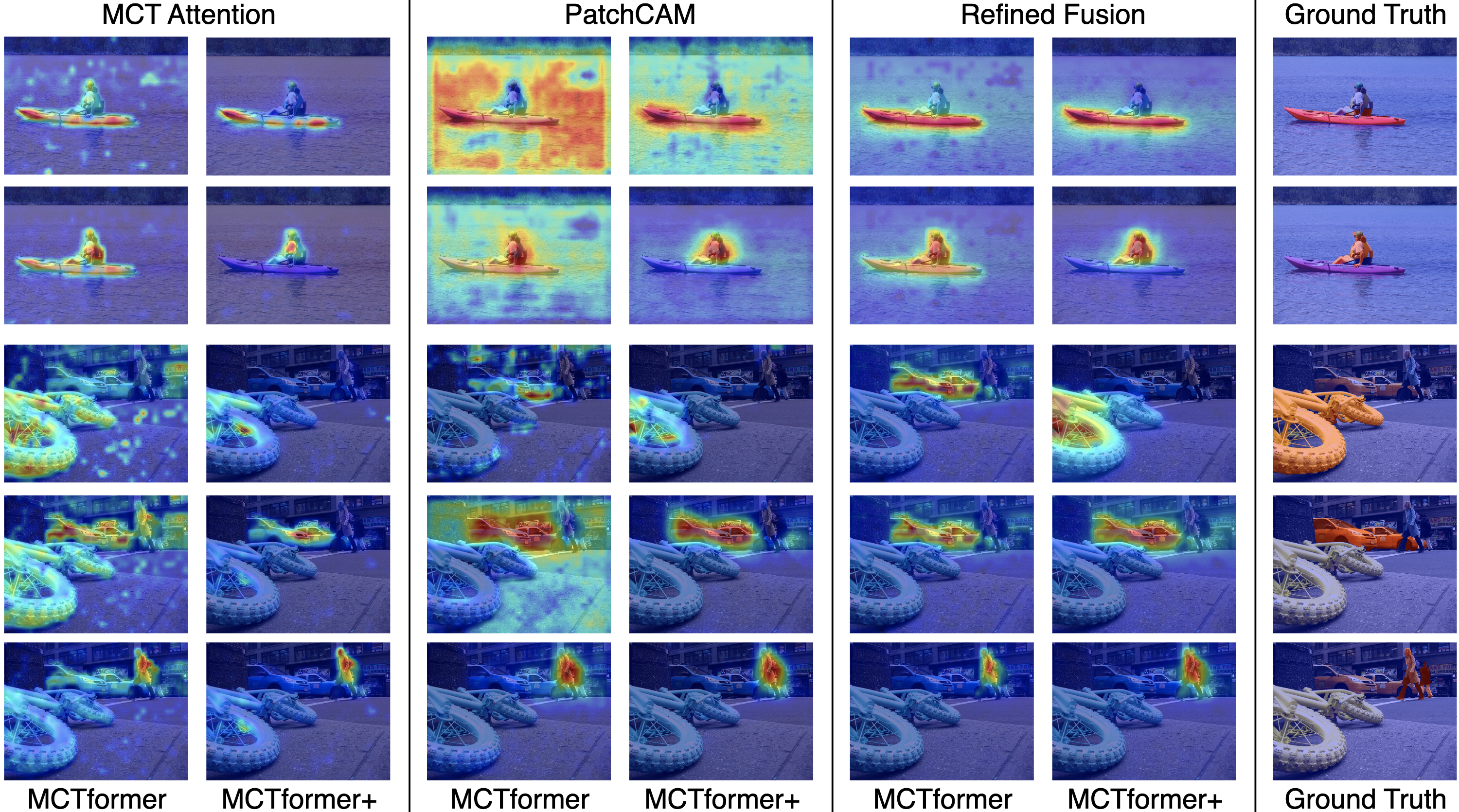}
\vspace{-8pt}
\end{center}
\caption{Qualitative comparison of the class-specific localization maps generated by the proposed MCTformer~\cite{xu2022multi} and MCTformer+.}
\label{fig:mct_plus_cam}
\vspace{-12pt}
\end{figure*}

\begin{table}[t]
\caption{Comparison of mIoUs (\%) for the seed maps generated by different configurations of the proposed MCTformer+ on PASCAL VOC \textit{train} set. }
\vspace{-8pt}
\label{tab:module_abla}
\small
\centering
\begin{tabular}{lcccc}
\toprule 
\multirow{2}{*}{} & \multirow{2}{*}{Pooling} & \multirow{2}{*}{CCT} & \multicolumn{2}{c}{Training image size} \\
                  &                          &                      & $224\times224$ & $448\times448$     \\
\toprule
\multirow{4}{*}{MCTformer+} & GAP& &63.2& 63.6 \\
&GAP&\checkmark&64.5&66.1\\
&GWRP& &64.8 &65.9\\
&GWRP&\checkmark&\textbf{65.8}&\textbf{68.8}\\
                            \bottomrule
\end{tabular}
\vspace{-4pt}
\end{table}
\begin{table}[t]
\caption{Comparison of mIoUs (\%) for the class-specific localization maps by using different global spatial pooling methods in the CAM module of the proposed \nn~on PASCAL VOC \textit{train} set. }
\vspace{-8pt}
\label{tab:GWRP}
\small
\centering
\begin{tabular}{lccc}
\toprule 
          & GMP & GAP & GWRP                   \\ \midrule
mIoU &59.4&66.1&\textbf{68.8}\\
  \bottomrule 
\end{tabular}
\vspace{-8pt}
\end{table}

\par\noindent\textbf{Effect of map refinement.} As reported in \Table{abla_mct_pgt} and \Table{abla_mct_seg}, the efficacy of the proposed PatchAffinity, \ie, the transformer attention between patch tokens, remains consistent in enhancing the generated class-specific localization maps and the resulting segmentation performances across different variants of the proposed \nn~with and without the CAM module.
Their resulting seed maps are shown to be consistently improved by significant margins. Their segmentation performances are boosted by gains of 3.2\% and 1.5\%, respectively. 
As presented in \Fig{mct_cam} (d) and (h), the refined class-specific localization maps exhibit enhanced completeness and smoother object contour. 
These results provide further evidence of the significant advantages of the proposed method in generating effective pair-wise affinity requiring no additional computations.

\par\noindent\textbf{Different class prediction methods.} 
In order to achieve effective class-specific token learning, we have evaluated different methods of processing output class tokens and investigated their effects on the resulting multi-class transformer attention.  
\Table{average-pool} shows that average pooling performs the best with an mIoU of 47.2\%, while max pooling has the lowest performance at 26.8\%. Utilizing a fully connected layer yields an mIoU of 41.5\%. This confirms our initial design motivation, as average pooling encourages class tokens to attend to more relevant patches, leading to improved spatial context for localization compared to max pooling, while including additional parameters through a fully-connected layer may complicate the model's learning for discriminative localization.

\par\noindent\textbf{Number of layers for attention fusion.} We evaluated the quality of the class-specific localization maps generated by combining the multi-class transformer attention maps from different layers of the proposed~\nn~without the CAM module.~Following~\cite{wang2020self}, we employed three evaluation metrics, \ie, false positive rate (FP), false negative rate (FN), and mIoU. 
FP serves as an indicator of the extent of the over-activation problem present in the CAM maps, while FN provides insight into the degree of the under-activation problem. 
\Fig{lineplot} shows that combining attention maps from more layers tends to produce over-activated class-specific localization maps. 
This suggests that shallow layers primarily contribute to learning generic low-level representations, which may not be advantageous for precise high-level semantic localization. 
On the other hand, reducing the number of layers leads to more discriminative object localization maps but at the expense of reduced activation coverage.
\Fig{lineplot} shows that fusing the attention maps from the last three layers yields the best-quality seeds
(mIoU of 47.2\%). 

\subsection{Ablation Studies on \nn+}

\par\noindent\textbf{Effect of different global spatial pooling methods.} Table~\ref{tab:GWRP} shows the dense localization results by learning with different spatial pooling methods to aggregate patch tokens for class predictions. Global Max Pooling (GMP) aggregates the feature maps by taking the maximum activation value across spatial dimensions (\textit{i.e.,} height and width) and it leads to a dense localization result of 59.4\% in mIoU. GMP helps in localizing the most relevant and discriminative information for each class prediction. However, it can lead to incomplete object localization maps. Global Average Pooling (GAP) aggregates feature maps by taking the average value across spatial dimensions. GAP treats all spatial locations equally, which implicitly assumes that all spatial locations have an equal probability of containing useful information such as objects of interest. This contributes to generating more comprehensive object localization maps with a significantly improved mIoU of 66.1\%, which is however prone to cover irrelevant information. In contrast, GWRP performs weighted pooling where the weight assigned to each spatial position is determined by its corresponding rank. Typically, the lower the rank, the higher the weight assigned to that spatial position. GWRP strikes a balance between considering the spatial distribution (like GAP) and highlighting salient features (like GMP) through rank-based weighting, leading to the best localization mIoU of 68.8\%. As depicted in~\Table{module_abla}, the efficacy of GWRP remains consistent in the proposed \nn+ when using different training resolutions. As shown in \Fig{mct_plus_cam}, by using GWRP, \nn+ generates more accurate PatchCAM maps with significantly reduced background noises compared to \nn~with GAP, indicating the effectiveness of GWRP in guiding the model to only focus on object regions.

\begin{figure}
\begin{center}
\includegraphics[width=.4\textwidth]{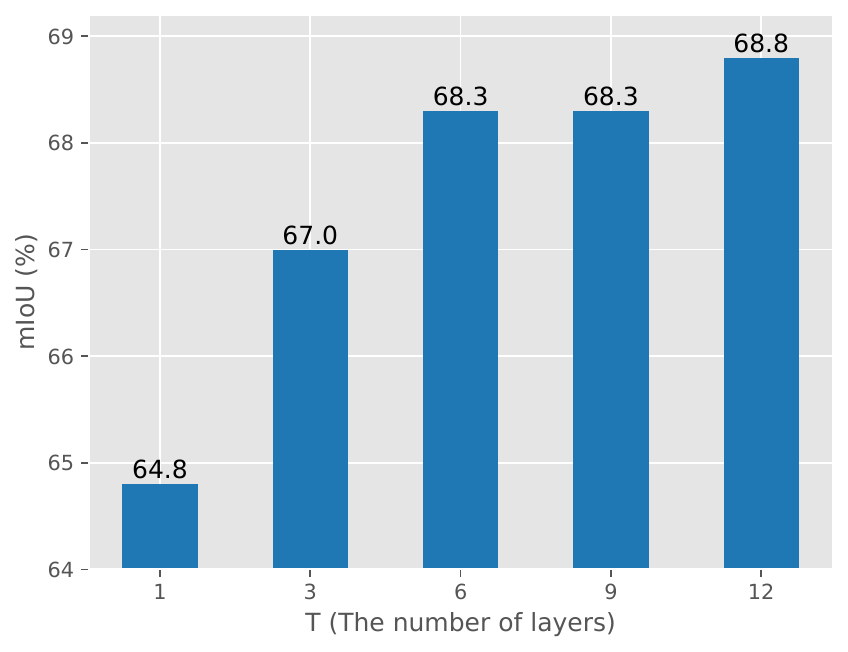}
\end{center}
\vspace{-8pt}
\caption{Comparison of mIoUs (\%) for the class-specific localization maps led by applying the proposed CCT module in different layers of the proposed \nn+ on PASCAL VOC \textit{train} set. The x-axis T denotes the number of layers (from top to bottom) where CCT is applied.}
\label{fig:barplot}
\vspace{-8pt}
\end{figure}

\par\noindent\textbf{Effect of different layers constrained by CCT.} We investigated the effect of applying the proposed CCT module on the output class tokens from different layers. 
\Fig{barplot} reveals that solely applying CCT on the output class tokens of the last layer yields the worst localization mIoU of 64.8\%, which shows a degradation in performance compared to not using CCT at all. The reason behind this lies in the limited influence the loss gradients have on the early layers and the presence of two losses on the final output class tokens poses challenges for optimization. 
In contrast, applying CCT on the output class tokens from all layers leads to the best localization performance, with an mIoU of 68.8\%. Since the input class tokens share the same initialization, without any additional guidance, the output class tokens from early layers tend to become similar. This similarity hampers the generation of diverse class tokens by the middle or top layers, given the interdependence among layers. This indicates that incorporating dense constraints on all layers significantly enhances the learning of discriminative class tokens. As depicted in~\Table{module_abla}, the efficacy of the proposed CCT remains consistent when using different training resolutions. As shown in \Fig{mct_plus_cam}, by using the CCT modules, \nn+ is shown to produce more class-discriminative localization maps in both MCT attention and PatchCAM, compared to \nn. For instance, different class tokens of \nn~for ``person" and ``boat" both attend to the patches belonging to the ``boat" class in the transformer attention, while different class tokens of \nn+ are shown to correctly attend to different patches.

\section{Conclusions}
\label{sec:conclusion}
In this paper, we introduce \nn+, an innovative transformer-based framework, designed to generate precise object localization maps tailored to specific classes. We demonstrate that learning multiple class tokens enables the discovery of class-specific localization information from class-to-patch attention. The proposed Contrastive-Class-Token module further facilitates the generation of more class-discriminative localization maps. Moreover, the patch-to-patch attention learns pairwise affinities, which can effectively refine the localization maps to be more accurate. Additionally, we show the seamless integration of our proposed framework with the CAM mechanism, jointly contributing to high-quality pseudo ground-truth labels for WSSS. Overall, our findings highlight the efficacy and simplicity of the proposed MCTformer+, along with its potential to significantly enhance the performance of WSSS models. Our approach surpasses existing methods and establishes new benchmarks in the field of WSSS. 

\bibliographystyle{IEEEtran}
\bibliography{mybib}


\begin{IEEEbiography}[{\includegraphics[width=1in,height=1.25in,clip,keepaspectratio]{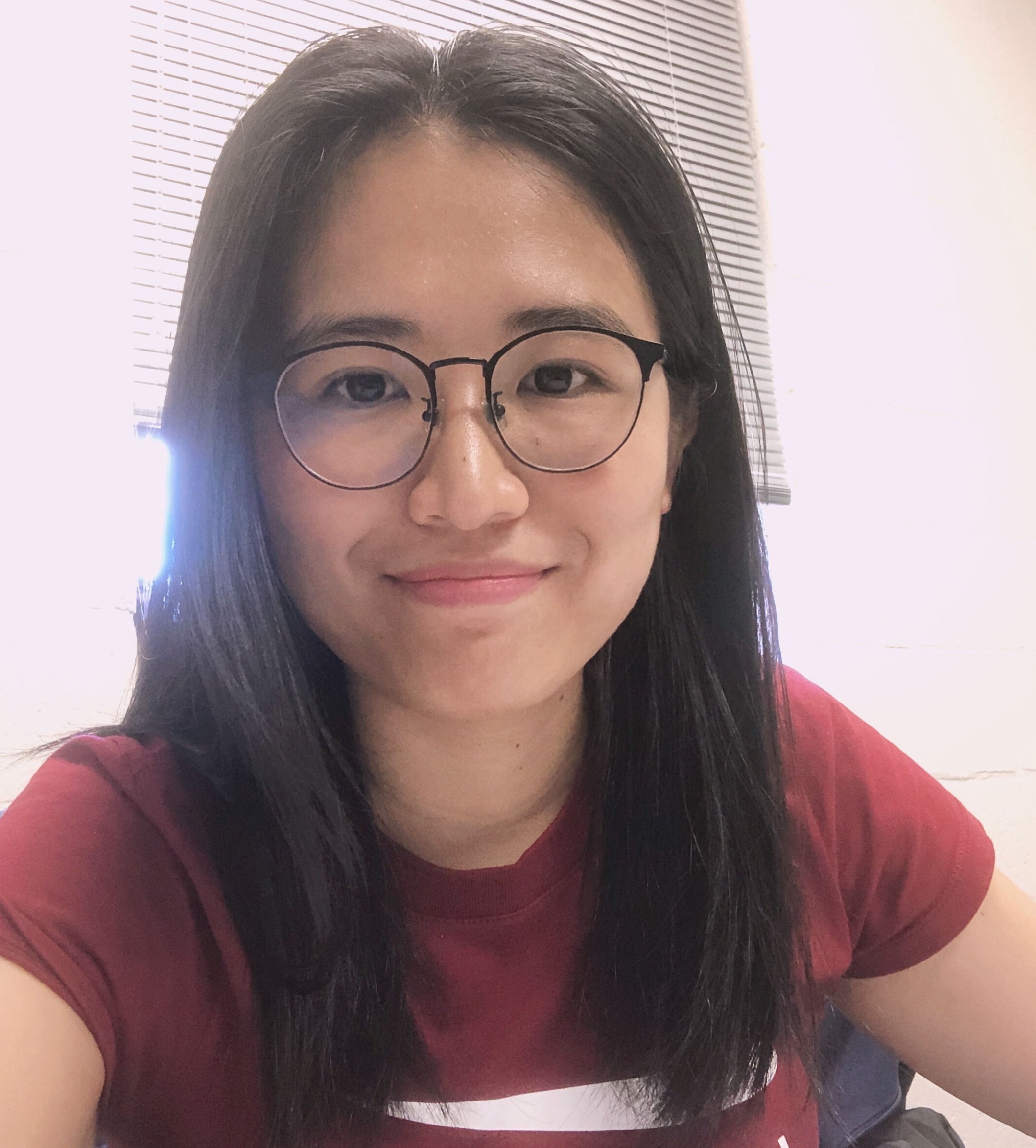}}]{Lian Xu}
received the Ph.D. degree from the Department of Computer Science and Software Engineering at the University of Western Australia. She is currently a research fellow at the University of Western
Australia. Her research interests include computer vision and machine learning.
\end{IEEEbiography}

\begin{IEEEbiography}[{\includegraphics[width=1in,height=1.25in,clip,keepaspectratio]{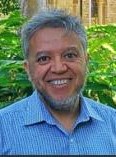}}]{Mohammed Bennamoun}
is Winthrop Professor in the Department of Computer Science and Software Engineering at UWA and is a researcher in computer vision, machine/deep learning, robotics, and signal/speech processing. He has published 4 books
(available on Amazon), 1 edited book, 1 Encyclopedia article (by invitation), 14 book chapters, 160+ journal papers, 250+ conference publications, 16 invited \& keynote publications. His h-index is 60 and his number of citations is close to 16,000 (Google Scholar). He was awarded 65+ competitive research grants (approx. \$20+ million in funding) from the Australian Research Council, and numerous other Government, UWA and industry Research Grants.
He has delivered conference tutorials at major conferences, including: IEEE Computer Vision and Pattern Recognition (CVPR 2016), Interspeech 2014, IEEE International Conference on Acoustics Speech and Signal Processing (ICASSP) and European Conference on Computer Vision (ECCV). He was also invited to give a Tutorial at an International Summer School on Deep Learning (DeepLearn 2017. He widely collaborated with researchers from within Australia (e.g. CSIRO), and internationally (e.g. Germany, France, Finland, USA). He served for two terms (3 years each term) on the Australian Research Council (ARC) College of Experts, and the ARC ERA 2018 (Excellence in Research for Australia).
\end{IEEEbiography}

\begin{IEEEbiography}[{\includegraphics[width=1in,height=1.25in,clip,keepaspectratio]{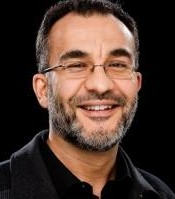}}]{Farid Boussaid} received the M.S. and Ph.D. degrees in microelectronics from the
National Institute of Applied Science (INSA),Toulouse, France, in 1996 and 1999 respectively. He joined Edith Cowan University, Perth, Australia, as a Postdoctoral Research Fellow, and a Member of the Visual Information Processing Research Group in 2000. He joined the University of Western Australia, Crawley, Australia, in 2005, where he is currently a Professor. His current research interests include neuromorphic engineering, smart sensors, and machine learning.
\end{IEEEbiography}

\begin{IEEEbiography}
[{\includegraphics[width=1in,height=1.25in,clip,keepaspectratio]{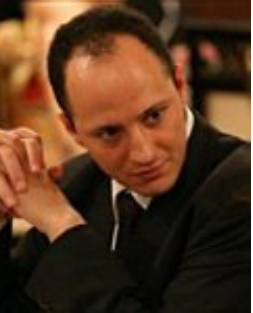}}]{Hamid Laga}  is  a  Professor at Murdoch University (Australia). His research interests span various fields of machine learning, computer vision, computer graphics, and pattern recognition, with a special focus on the 3D reconstruction, modeling, and analysis of static and deformable 3D objects, and on machine learning for agriculture and health. He is the recipient of  the Best Paper Awards at SGP2017, DICTA2012, and SMI2006.
\end{IEEEbiography}

\begin{IEEEbiography}[{\includegraphics[width=1in,height=1.25in,clip,keepaspectratio]{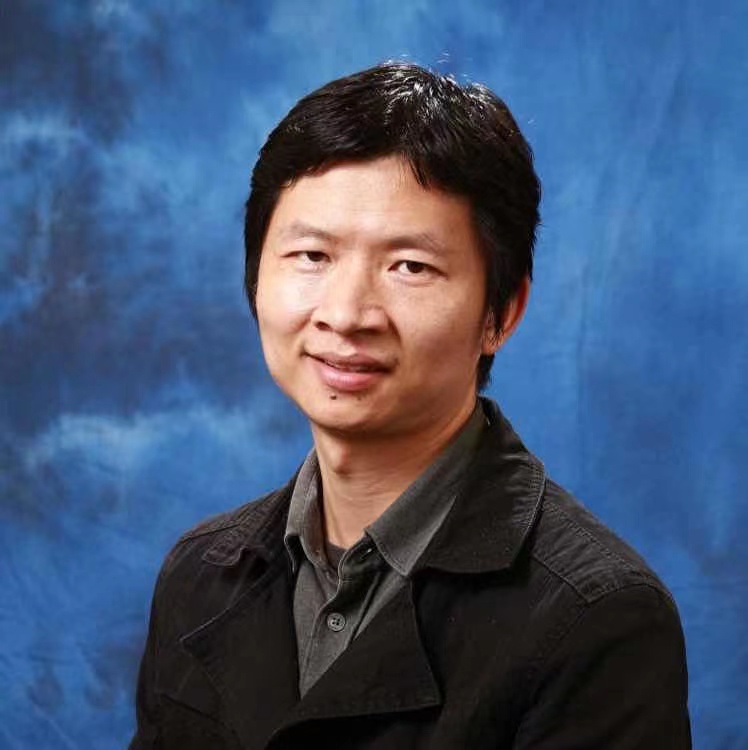}}]{Wanli Ouyang} received
the Ph.D. degree from the Department of Electronic Engineering, The Chinese University of Hong Kong, Hong Kong, in 2010. He is currently a Professor with the Shanghai AI Laboratory, Shanghai, China. His research interests include image processing, computer vision, and pattern recognition.
\end{IEEEbiography}

\begin{IEEEbiography}
[{\includegraphics[width=1in,height=1.25in,clip,keepaspectratio]{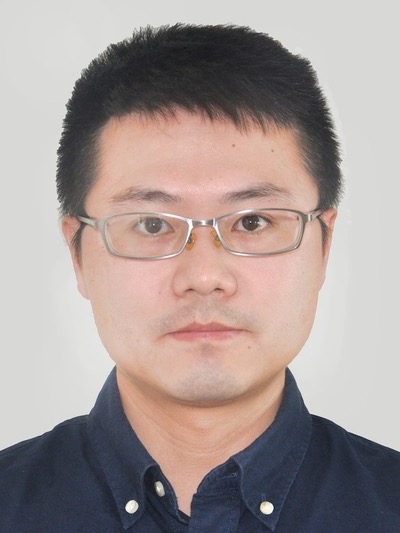}}]
{Dan Xu} is an Assistant Professor in the Department of Computer Science and Engineering at HKUST. He was a Postdoctoral Research Fellow
in VGG at the University of Oxford. He was a
Ph.D. in the Department of Computer Science at
the University of Trento. He was also a research
assistant of MM Lab at the Chinese University of
Hong Kong. He received the best scientific paper
award at ICPR 2016, and a Best Paper Nominee
at ACM MM 2018. He served as Area Chairs at multiple main-stream conferences including CVPR, AAAI, ACM Multimedia, WACV, ACCV and ICPR.
\end{IEEEbiography}






\end{document}